\documentclass[10pt,twocolumn,letterpaper]{article}

\usepackage{cvpr}
\usepackage{times}
\usepackage{epsfig}
\usepackage{graphicx}
\usepackage{amsmath}
\usepackage{amssymb}
\usepackage{mathtools}
\usepackage{caption}
\usepackage{subcaption}
\usepackage{multirow}
\usepackage{enumitem}
\usepackage{color}
\usepackage{xcolor}
\def\red#1{\textcolor[rgb]{1,0,0}{#1}}

\newcommand{\doublecheck}[1]{\textcolor{black}{#1}}
\newcommand{\keypoint}[1]{\vspace{0.1cm}\noindent\textbf{#1}\quad}
\newcommand{\cut}[1]{}
\usepackage{algorithm}
\usepackage{algpseudocode}

\usepackage[nopar]{lipsum}
\usepackage{arydshln}
\def\red#1{\textcolor[rgb]{1,0,0}{#1}}
\def\blue#1{\textcolor[rgb]{0,0,1}{#1}}

\usepackage[nolist]{acronym} 

\makeatother
\usepackage[pagebackref,breaklinks,colorlinks]{hyperref}

\cvprfinalcopy 

\def\cvprPaperID{1409} 

\ifcvprfinal\fi
\begin{document}

\begin{acronym}[GloVeeee] 
\acro{DUC}{Document Understanding Conferences}
\acro{GPU}{graphics processing unit}
\acro{GloVe}{global vectors for word representation}
\acro{LSTM}{Long Short-term Memory}
\acro{AMS}{automatic multi-modal summarization}
\acro{IR}{information retrieval}
\acro{IE}{information extraction}
\acro{QA}{question answering}
\acro{AI}{artificial intelligence}
\acro{RNN}{recurrent neural network}
\acro{SIFT}{scale invariant feature transform}
\acro{MFCC}{Mel-frequency cepstral coefficients}
\acro{NLP}{natural language processing}
\acro{SVD}{Singular Value Decomposition}
\acro{SVM}{Support Vector Machine}
\acro{MLE}{maximum likelihood estimation}
\acro{BERT}{Bidirectional Encoder Representations from Transformers}
\acro{GRU}{Gated Recurrent Unit}
\acro{BLEU}{Bilingual evaluation understudy}
\acro{ROUGE}{Recall-Oriented Understudy for Gisting Evaluation}
\acro{HTML}{Hypertext Markup Language}
\acro{MMSS}{multi-modal sentence summarization}
\acro{CPU}{central processing unit}
\acro{NMT}{neural machine translation}
\acro{LSA}{least common ancestor}
\acro{LCS}{longest common sub-sequence}
\acro{BP}{Brevity Penalty}
\acro{SNLI}{Stanford Natural Language Inference}
\acro{LTS}{Long Term Support}
\acro{RAM}{Random Access Memory}
\acro{GUI}{graphical user interface}
\acro{SBD}{sentence boundary detection}
\acro{OOV}{out of vocabulary}
\acro{FG-SBIR}{fine-grained sketch-based image retrieval}
\acro{RL}{reinforcement learning}
\acro{SBIR}{sketch-based image retrieval}
\end{acronym}
 
\title{Sketching \textit{without} Worrying: Noise-Tolerant Sketch-Based Image Retrieval}
\author{Ayan Kumar Bhunia\textsuperscript{1} \hspace{.2cm}  \hspace{.2cm}  Subhadeep Koley\textsuperscript{1,2} \hspace{.2cm}  Abdullah Faiz Ur Rahman Khilji\thanks{Interned with SketchX} \hspace{.2cm} Aneeshan Sain\textsuperscript{1,2} \\  Pinaki nath Chowdhury \textsuperscript{1,2} \hspace{.2cm} 
Tao Xiang\textsuperscript{1,2}\hspace{.2cm}  Yi-Zhe Song\textsuperscript{1,2} \\
\textsuperscript{1}SketchX, CVSSP, University of Surrey, United Kingdom.  \\
\textsuperscript{2}iFlyTek-Surrey Joint Research Centre on Artificial Intelligence.\\
{\tt\small \{a.bhunia, s.koley, a.sain, p.chowdhury, t.xiang, y.song\}@surrey.ac.uk} 
}

\maketitle
\ifcvprfinal\thispagestyle{empty}\fi

\begin{abstract}
\vspace{-0.2cm}
Sketching enables many exciting applications, notably, image retrieval. The fear-to-sketch problem (i.e., ``I can't sketch") has however proven to be fatal for its widespread adoption. This paper tackles this ``fear" head on, and for the first time, proposes an auxiliary module for existing retrieval models that predominantly lets the users sketch without having to worry. We first conducted a pilot study that revealed the secret lies in the existence of noisy strokes, but not so much of the ``I can't sketch". We consequently design a stroke subset selector that {detects noisy strokes, leaving only those} which make a positive contribution towards successful retrieval. Our Reinforcement Learning based formulation quantifies the importance of each stroke present in a given subset, based on the extent to which that stroke contributes to retrieval. When combined with pre-trained retrieval models as a pre-processing module, we achieve a significant gain of 8\%-10\% over standard baselines and in turn report new state-of-the-art performance. Last but not least, we demonstrate the selector once trained, can also be used in a plug-and-play manner to empower various sketch applications in ways that were not previously possible.

\end{abstract}






\vspace{-0.6cm}
\section{Introduction}
\vspace{-0.1cm}
Thanks to the convenience of interactive touchscreen devices, sketch-based image retrieval (SBIR) \cite{collomosse2019livesketch, dey2019doodle, dutta2019semantically, Sketch3T} has emerged as a practical means of image research that is complementary to the conventional text-based retrieval \cite{lee2018stacked}. Although initially developed for a category-level setting \cite{shen2018zero, ribeiro2020sketchformer, zhang2018generative}, of late SBIR has undertaken a \textit{fine-grained} shift to better reflect the inherent fine-grained characteristics (pose, appearance detail, etc) of sketches \cite{song2017deep, yu2016sketch, bhunia2020sketch}.

Despite great strides made \cite{bhunia2021more, pang2019generalising, PartialSBIR}, the \textit{fear-to-sketch} has proven to be fatal for its omnipresence -- a \emph{``I can't sketch"} reply is often the end of it. This \emph{``fear"} is predominant for fine-grained SBIR (FG-SBIR), where the system dictates users to produce even more faithful and diligent sketches than that required for category-level retrieval \cite{collomosse2019livesketch}. 

In this paper, we tackle this \emph{``fear"} head-on and propose for the first time a \emph{pre-processing} module for FG-SBIR that essentially let the users sketch without the worry of \emph{``I can't''}. We first experimentally show that, in most cases it is not about how bad a sketch is -- most \textit{can} sketch (even a rough outline) -- the devil lies in the fact that users typically draw irrelevant (noisy) strokes that are detrimental to the overall retrieval performance (see Section \ref{sec3}). This observation has largely inspired us to alleviate the \emph{``can't sketch"} problem by \emph{eliminating} the noisy strokes through selecting an optimal subset that \textit{can} lead to effective retrieval. 

\begin{figure}[t]
\centering
\includegraphics[height=3.5cm,width=\linewidth]{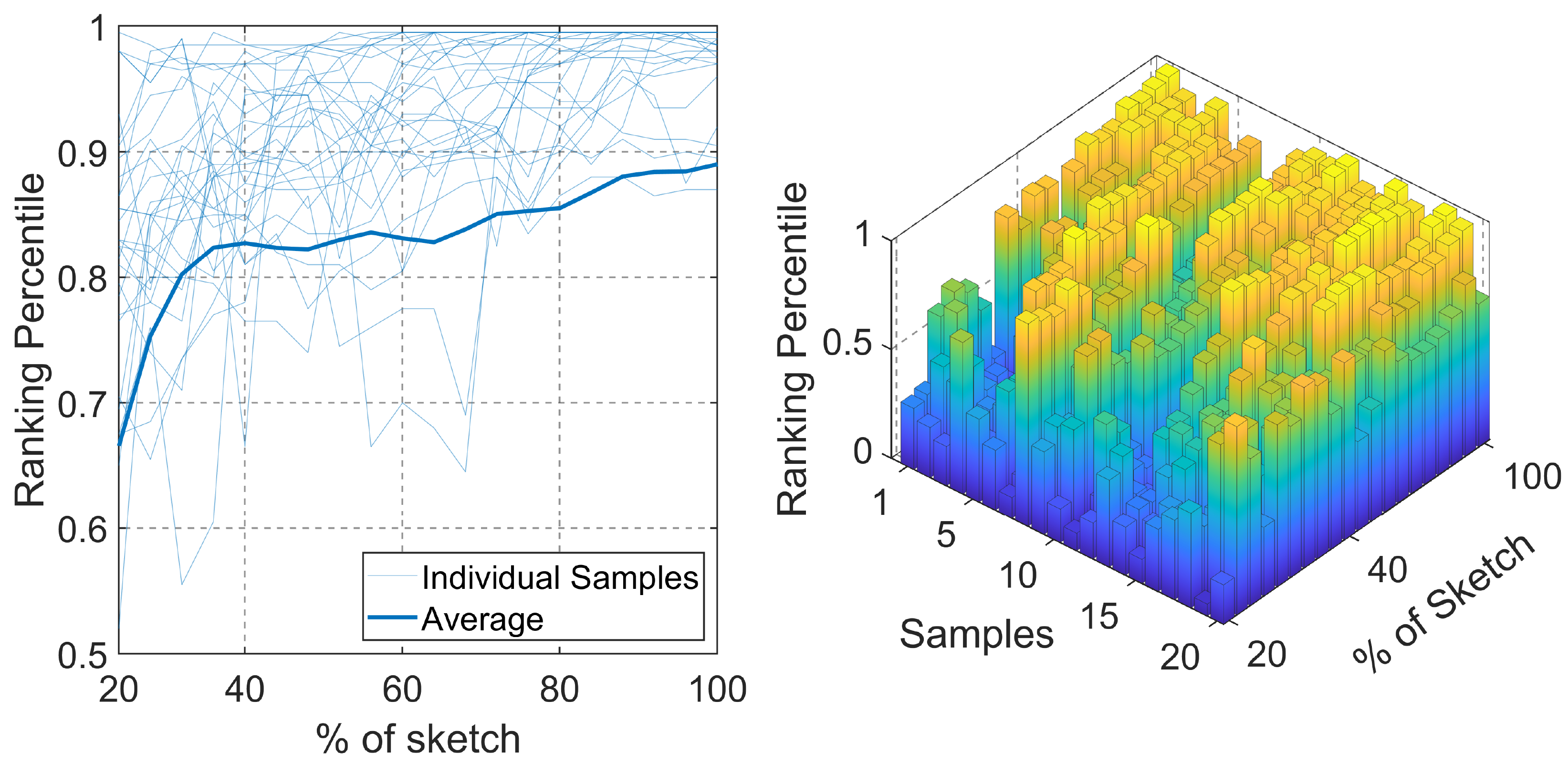}
\vspace{-0.7cm}
\caption{(a) While the \emph{average} ranking percentile increases as the sketching proceeds from starting towards completion, \emph{unwanted sudden drops} have been noticed for many individual sketches due to noisy/irrelevant strokes drawn. (b) The same thing is visualised with number of samples in the third axis to get an overall statistics on QMUL-Shoe-V2 dataset.}
\label{fig1}
\vspace{-0.6cm}
\end{figure}

This problem might sound trivial enough -- e.g., how about considering all possible stroke subsets as training samples to gain model invariance against noisy strokes? Albeit theoretically possible, the highly complex nature of this process (i.e., $\mathcal{O}(2^N)$) quickly renders this naive solution infeasible, especially when the number of strokes in free-hand sketches can range from an average of $N={9}$ to a max of $N={15}$ in fine-grained SBIR datasets (QMUL-ShoeV2/ChairV2 \cite{yu2016sketch, song2017deep}). Most importantly, augmenting the training data by random stroke dropping would lead to a noisy gradient during training. This is because out of all possible subsets, many of these augmented sketch subsets are too coarse/incomplete to convey any meaningful information to represent the paired photo. Therefore, instead of naively learning the invariance, we advocate for finding meaningful subsets that can sustain efficient retrieval.


Our solution generally rests with detecting noisy strokes and leaving only those that positively contribute to successful retrieval. {We achieve that by proposing a mechanism to quantify the \emph{importance} of each stroke present in a given stroke-set, based on the extent to which that stroke is \emph{worthy for retrieval} (i.e, makes a positive contribution).} We work on vector sketches\cite{bhunia2021vectorization} in order to utilise stroke-level information, and propose a sketch stroke subset selector that learns to determine a binary action for every stroke -- whether to include that particular stroke to the query stroke subset, or not. The stroke subset selector is designed via a hierarchical Recurrent Neural Network (RNN) that models the compositional relationship among the strokes. Once the stroke subset is obtained, it is first \emph{rasterized} then passed through a pre-trained FG-SBIR model \cite{yu2016sketch} to obtain a ranking of target photos against the ground-truth photo. The main objective is to select a particular subset that will rank the paired ground-truth photo towards the top of the ranking list. We use Reinforcement Learning (RL) based training due to the non-differentiability of rasterization operation. As explicit stroke-level ground-truth for the optimal subset is absent, we seek to train our stroke-subset selector with the help of pre-trained FG-SBIR for reward computation.  \cut{ that exists between stroke subset selector and pre-trained FG-SBIR.} In particular, we use the actor-critic version of proximal policy optimisation (PPO) to train the stroke subset selector.


Apart from the main objective of noisy stroke elimination, the proposed method also enables a few secondary sketch applications (Section \ref{byproduct}) in a plug-and-play manner. First, we show that a pre-trained stroke selector can be used as a \emph{stroke importance quantifier} to guide users to produce a sketch ``just'' enough for successful retrieval. 
Second, we demonstrate that it can significantly speed up existing works on interactive ``on-the-fly'' retrieval \cite{bhunia2020sketch} removing the need for incomplete rasterized sketch to be unnecessarily passed for inference multiple times.
Third, besides benefiting FG-SBIR, our subset selector module can also act as a faithful \emph{sketch data augmenter} over random stroke dropping without much computational overhead. That is, instead of costly operation like sketch deformation \cite{yu2017sketch} or unfaithful approximation like edge/contour-map as soft ground-truths \cite{chen2018sketchygan}, users can effortlessly generate $n$ most representative subsets to augment training for many downstream tasks. \cut{(e.g., sketch generation \cite{ha2017neural}, vectorization \cite{das2020beziersketch}, combination of both \cite{das2021cloud2curve}).}

In summary our contributions are, (a) We tackle the fear-to-sketch problem for sketch-based image retrieval for the first time, (b) We formulate the ``can't sketch" problem as stroke subset selection problem following detailed experimental analysis, (c) We propose a RL-based framework for stroke subset selection that learns through interacting with a pre-trained retrieval model. (d) We demonstrate our pre-trained subset selector can empower other sketch applications in a plug-and-plug manner.


\vspace{-0.15cm}
\section{Related Works}
\vspace{-0.3cm}

\keypoint{Category-level SBIR:} Category-level SBIR aims at retrieving category-specific photos from user given query sketches. Like any other retrieval system, Deep Neural Networks have become a de-facto choice for any recent SBIR frameworks \cite{dutta2019semantically, dey2019doodle, ribeiro2020sketchformer, zhang2018generative, collomosse2019livesketch, DoodleIncremental} over early hand-engineered feature descriptors \cite{tolias2017asymmetric}. Overall, category level SBIR makes use of Siamese networks based on either CNN \cite{collomosse2019livesketch, dey2019doodle}, RNN \cite{xu2018sketchmate}, Transformer \cite{ribeiro2020sketchformer} or their combinations \cite{collomosse2019livesketch} along with a triplet-ranking objective to learn a joint embedding space. A distance metric is used to rank the gallery photos against the learned embedding space for a given query sketch for retrieval. Further efforts have been made through zero-shot SBIR \cite{dey2019doodle,yelamarthi2018zero} for cross-category generalisation, and employing binary hash-code embedding \cite{liu2017deep,shen2018zero} \cut{over continuous feature vectors} to reduce the computational complexity. 

\keypoint{Fine-grained SBIR:} 
Sketch holds a noteworthy advantage in its potential to depict fine-grained properties of the target image, which are hard to describe via other query mediums \cite{song2017fine} like text or attribute. Consequently, interest surged in fine-grained SBIR \cite{yu2016sketch}, which aims at \emph{instance-specific} matching for a user given query sketch. Initially starting with graph-matching models \cite{pang2019generalising}, FG-SBIR research gained traction with the advent of various deep-learning based approaches \cite{yu2016sketch, song2017deep, bhunia2020sketch, bhunia2021more}. Yu \etal \cite{yu2016sketch} first pioneered deep triplet-ranking based \emph{siamese networks} for learning a joint embedding space with instance-wise matching criteria. This was further augmented via attention with higher-order retrieval loss \cite{song2017deep}, cross-domain image generation \cite{pang2017cross}, text tags \cite{song2017fine}, etc. Recent FG-SBIR works include advanced methods like hierarchical co-attention \cite{sain2020cross}, reinforcement learning-based early retrieval \cite{bhunia2020sketch}, semi-supervised generation-retrieval joint training \cite{bhunia2021more}, etc.

While sketches are significantly subjective to user's style \cite{sain2021stylemeup} and vary considerably depending on the drawer's drawing skill \cite{bhunia2020sketch}, these earlier works \emph{assumed} the existing annotated fine-grained dataset to be \emph{perfect}. In other words,  a \emph{rigid} assumption is made that every annotated paired sketch is a perfect depiction of the paired photo. In this work, we argue that \emph{`all sketches are sketchy'}, which holds stronger significance for fine-grained SBIR, as every stroke of annotated sketch \cite{yu2021fine} represents a specific part of the paired photo, and the free-flow nature of amateur sketching is likely to introduce noise no matter how carefully it is drawn.

\keypoint{Modelling Partial Sketches:} ``Sketch'' being an interactive medium, is drawn sequentially in a stroke-by-stroke manner. Moreover, due to its subjective nature, \cut{ it can attain varying level of precision and} the same sketch might be perceived as partial or complete based on the user's perception. Users can retrieve photos \cite{bhunia2020sketch}, create \cite{wang2021sketch} imaginative visual-art, or edit existing photos \cite{jo2019sc} through repeated interactions with the AI agent. Therefore, on-the-fly interaction with sketches requires sketch-based models to be capable of handling partial sketches. For instance, Sketch-RNN \cite{ha2017neural} can predict probable final sketch endings using a variational autoencoder trained on the vector sketch coordinates. Furthermore, attempts have been made to directly recognise partial sketches \cite{liu2019sketchgan} and achieve sketch-to-photo generation \cite{ghosh2019interactive} from incomplete sketch input, where both works involve  a sketch-completion module based on image-to-image translation. Recently, on-the-fly FG-SBIR \cite{bhunia2020sketch} has been introduced to retrieve even from a few elementary strokes as soon as the users start drawing. Overall, these works try to include random synthetic partial sketches during training to achieve their respective goals, but here we aim to answer \emph{``whether a partial sketch has sufficient representative information/discriminative potential to retrieve photos faithfully''}. Furthermore, we aim to quantify the instant at which a sequentially drawn sketch would reach the optimum threshold point where it is representative enough for downstream tasks (e.g., retrieval). By doing so, we can faithfully train models with sufficiently representative partial sketches instead of randomly dropping strokes and ignoring instances where the synthetic partial sketch is too coarse to convey any meaning.    
 
\keypoint{Reinforcement Learning in Vision:} Reinforcement Learning (RL) \cite{kaelbling1996reinforcement} has been applied in different vision problems \cite{liang2017deep, wang2019reinforced}. 
RL becomes handy when there exists a non-differentiable way to quantify the \textit{goodness} of the network's state unlike differentiable loss function with hard-labels. Instead, learning progresses via interactions \cite{duong2019automatic, han2019deep} with the environment. Particularly in sketch community, RL has been leveraged for modelling sketch abstraction \cite{muhammad2018learning,muhammad2019goal}, retrieval \cite{bhunia2020sketch, bhunia2021more}, and  designing competitive sketching agent \cite{bhunia2020pixelor}. Here, our objective is to engage an RL agent to get rid of noisy sketch strokes for better retrieval.

\keypoint{Learning from noisy labels:} Despite significant progress from the community-generated labelled data, accurate labelling is challenging even for experienced domain experts \cite{song2021learning}. Therefore, a separate topic of study \cite{song2021learning,zhou2021robust,zhou2021robust} emerged, which aims at learning robust models even from the noisy data distribution. 
While the existing works \cite{han2018masking,tanno20219learning} mainly consider having access to a large, noisy dataset as well as a subset of carefully cleaned data for validation , our situation is even more difficult than usual. We assume that every annotated sketch is not an absolutely perfect matching sketch of the paired photo. Therefore, we aim to develop a noise-tolerant framework for FG-SBIR.







 

\vspace{-0.2cm}
\section{Pilot Study: What's Wrong with FG-SBIR?}
\vspace{-0.2cm}
\label{sec3}

\keypoint{Baseline FG-SBIR:} Instead of complicated pre-training \cite{pang2020solving} or joint-training \cite{bhunia2021more}, we use a three branch state-of-the-art Siamese network \cite{bhunia2021more} as our baseline retrieval model, which is considered to be a strong baseline till date. Each branch starts from ImageNet pre-trained VGG-16 \cite{simonyan2015very}, sharing equal weights. Given an input image $I \in \mathbb{R}^{H \times W \times 3}$, we extract the convolutional feature-map $\mathcal{F}({I})$, which upon global average pooling followed by  $l_2$ normalisation generates a $d$ dimensional feature embedding. This model has been trained with an anchor sketch (a), a positive (p) photo, and a negative (n) photo triplets $\{\bar{a},\bar{p},\bar{n}\}$ using \textit{triplet-loss} \cite{weinberger2009distance}. Triplet-loss aims at increasing the distance between anchor sketch and negative photo $\delta^{-}={||\mathcal F(\bar{a})-\mathcal F(\bar{n})||}_2$, while simultaneously decreasing the same between anchor sketch and positive photo $\delta^{+}={||\mathcal F(\bar{a})-\mathcal F(\bar{p})||}_2$. Therefore, the triplet-loss with margin $\mu>0$ can be written as:
\vspace{-0.2cm}
\begin{equation}
\label{eq1}
    \mathcal{L}_{Triplet}=max\{0, \delta^{+}-\delta^{-}+\mu\}
\vspace{-0.2cm}
\end{equation}
\vspace{-0.6cm}

\keypoint{Dual representation of sketch:} Recent study has emphasised on the dual representation \cite{bhunia2021vectorization} of sketch for self-supervised feature learning. In rasterized pixel modality $\mathcal{I}$, sketch can be represented as spatially extended image of size $\mathbb{R}^{H\times W\times 3}$. On the other side, in vector modality $\mathcal{V}$, the same sketch can be characterised by a sequence of strokes $(s_1, s_2, \cdots, s_K)$ where each stroke is a sequence of successive points $s_i = (v_1^i, v_2^i, \cdots, v_{N_i}^i)$, and each point is represented by an absolute 2D coordinate $v_n^i = (x_n^i, y_n^i)$ in a $H\times W$  canvas. Here, $K$ is number of strokes and $N_i$ is the number of points inside $i^{th}$ stroke. Individual strokes arise due to pen up/down \cite{ha2017neural} movement. 
Although sketch vectors can easily be recorded through touch screen-devices, generation of the corresponding rasterized sketch image needs a costly \cite{xu2021multigraph} \emph{rasterization} operation  $\mathcal{R}: \mathcal{V} \rightarrow \mathcal{I}$. Either modality, raster or vector,  has its own merits and demerits \cite{bhunia2021vectorization}. Apart from being more computationally efficient \cite{xu2021multigraph} than raster domain, vector modality also contains the stroke-by-stroke temporal information \cite{ha2017neural}. Nonetheless, sketch vectors lack the spatial information \cite{bhunia2021vectorization}  which is critical to model the fine-grained details \cite{bhunia2021more, bhunia2020sketch}. Consequently, rasterized sketch image is the standard choice \cite{pang2020solving, sain2021stylemeup, sain2020cross, yu2016sketch} for FG-SBIR despite having a higher computational overhead and lacking temporal information.

\keypoint{Preliminary analysis:} The performance barrier due to irrelevant strokes gets noticed under on-the-fly FG-SBIR \cite{bhunia2020sketch} setup. Instead of only evaluating the complete sketch, we start rendering  at the end of every new $k^{th}$ stroke drawn as the rasterized sketch image $S^{\mathcal{I}}_{k} = \mathcal{R}([s_1, s_2, \cdots, s_k])$ where $k=\{1,2, \cdots, K\}$, and pass it through the \emph{pre-trained} baseline FG-SBIR model to get the feature representation $\mathcal{F}(S^{\mathcal{I}}_{k})$, followed by ranking the gallery images against it. We make these following observations on Shoe-V2 \cite{yu2016sketch} dataset \emph{(Linear Limit)}: (i) As the sketch proceeds towards completion, the rank is supposed to be improved, however, we notice some unexpected dips in the performance in the later part of the drawing episode. This signifies that the later irrelevant strokes play a detrimental role, thereby degrading the retrieval performance (Fig.~\ref{fig1}). (ii) Compared to top@1(top@5) accuracy of $33.43\%(67.81\%)$ on using complete sketch for retrieval, if we consider best rank achieved at any of the instant during the sketch drawing episode as the retrieved result, top@1(top@5) accuracy extends to $42.54\%(73.28\%)$. (iii) Further, we note that the percentage of instances where subsequently added strokes drops the performance compared to the previous version $S^{\mathcal{I}}_{k}$ of the same sketch is $43.44\%$, which is a critical number.

\keypoint{Ablation on upper limit:} 
Prior analysis unfolds the necessity of dealing with irrelevant stroke, and we \emph{hypothesise that in many cases a subset of the strokes $K' \leq K$ could better retrieve the paired photo by excluding the irrelevant ones}. Different people follow varying stroke order for sketching. Therefore, in order to simulate different possible stroke orders and to estimate the upper limit that we can achieve through the smart stroke-subset selector, we do the following study. Given $K$ strokes in a sketch, we form $(2^{K}-1)$ stroke subsets taking \emph{any} number of strokes at a time. Unlike the ``on-the-fly’’ \cite{bhunia2020sketch} protocol,  this setting does not stick to a pre-recorded sequential order, rather it aims to find if there exists \emph{any subset} that can retrieve the paired photo better than the entire sketch set. Under this setting, we achieve an exceptionally high top@1(top@5) accuracy of  $66.37\%(88.31\%)$. However, evaluating with every possible stroke combination during real-time inference is \emph{impractical}, and we do not have any explicit way to select one final result.  Therefore, in this work, we seek to build a \emph{smart stroke-subset selector} as a \emph{pre-processing} module which when plugged in before any pre-trained FG-SBIR model  \cite{yu2016sketch, song2017fine}, will aim to construct the most representative subset to improve the overall accuracy.


\vspace{-0.2cm}
\section{Noisy Stroke Tolerant FG-SBIR}
\vspace{-0.2cm}
\keypoint{Overview:} Our preliminary study motivates us to design a stroke-subset selector to eliminate the noisy strokes for FG-SBIR. While \emph{raster sketch image} is essential \cite{bhunia2021more} to model the fine-grained correspondence, the stroke-level sequential information is missing in raster modality. Therefore, taking advantage of the dual representation \cite{bhunia2021vectorization} of the sketch, we model the stroke subset selector on the sequential vector space. In summary, our noise-tolerant FG-SBIR consists of two following modules connected in cascade: (a) \emph{stroke-subset selector} as pre-processing module working in vector space and (b) \emph{pretrained FG-SBIR} $\mathcal{F}$ that uses rasterized version of predicted subset for final retrieval. 



\vspace{-0.1cm}
\subsection{Stroke Subset Selector}
\vspace{-0.1cm}
\keypoint{Model:} Given sketch-photo pair $(S,P)$, the sketch $S$ can be represented as both raster image $S_{{I}}$ and stroke-level sequential vector $S_{{V}}=(s_1, s_2, \cdots, s_K)$. We design a stroke-subset selector $\mathcal{X}(\cdot)$ that takes $S_{{V}}$ as input, and aim to predict an optimal subset $\overline{S}_{V} = \mathcal{X}( S_{V})$ with $K'$ strokes where $K' \leq K$. However, selecting the optimal subset of stroke is an ill-posed problem.  Firstly, there is \emph{no explicit label} which represents the optimal stroke subset. In fact, there might be many sub-sets which can lead to successful retrieval. Furthermore, annotating the optimal stroke-subsets for the whole training dataset via brute-force iteration is computationally impractical \cite{bhunia2020pixelor}.

In our framework, we treat stroke subset selector as a binary categorical classification problem.  In other words, for a sketch of K strokes, we get an output of size $\mathbb{R}^{K\times2}$, where every row is softmax normalised and it represents a probability distribution $p(a_i|s_i)$ over two classes: $a \in \{\texttt{select}, \texttt{ignore}\}$. However, we do not have any explicit one-hot labels for this binary classification task. Therefore, we let the stroke sub-set selector agent to interact with the pre-trained FG-SBIR model, and $\mathcal{X}$ is learned  using a pre-trained FG-SBIR model $\mathcal{F}$ as a \emph{critic} which provides the training signal to $\mathcal{X}$.


\vspace{-0.4cm}
 \begin{figure}[!hbt]
\includegraphics[width=1.0\linewidth]{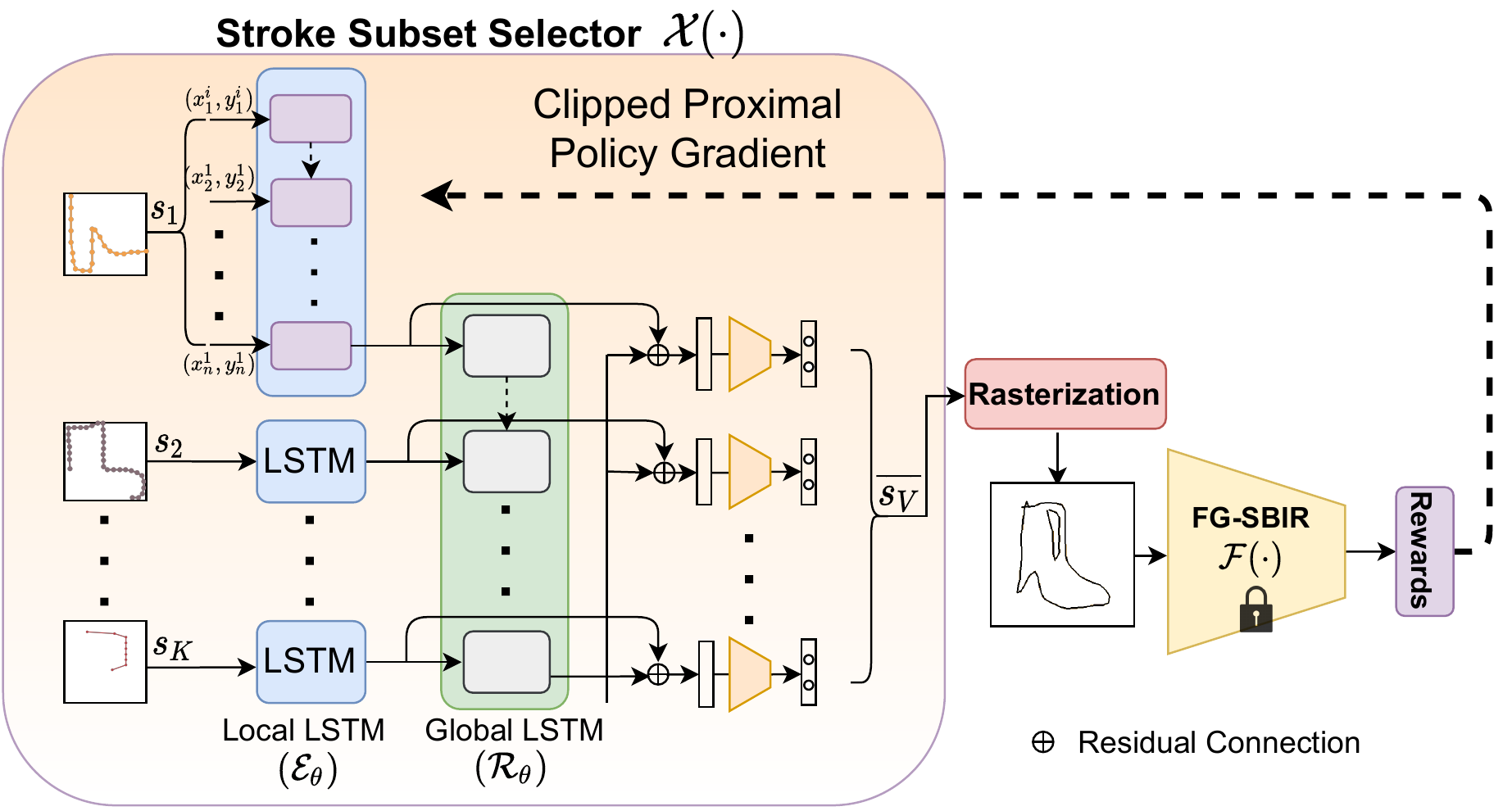}
\vspace{-0.7cm}
\caption{Illustration of Noise Tolerant FG-SBIR framework. Stroke Subset Selector $\mathcal{X(\cdot)}$ acts as a pre-processing module in the sketch vector space to eliminate the noisy strokes. Selected stroke subset is then rasterized and fed through an existing pre-trained FG-SBIR model for reward calculation, which is optimised by Proximal Policy Optimisation. For brevity, actor-only version is shown here. }
\vspace{-0.4cm}
\label{fig_archi}
\end{figure}

\keypoint{Architecture:} To design the architecture of stroke-level selector, we aim at preserving localised stroke-level information, as well as the compositional relationship \cite{aksan2020cose} among the strokes, which  together conveys the overall semantic meaning. Therefore, we employ a two-level hierarchical model comprising of a local stroke-embedding network ($\mathcal{E}_\theta$) and global relational network ($\mathcal{R}_\theta$)  to enrich each stroke-level feature about the global semantics. In particular, we feed individual stroke of size $\mathbb{R}^{N_i\times2}$ having $N_i$ points though a local stroke-embedding network $\mathcal{E}_\theta$ (e.g. RNN, LSTM or Transformer) whose weights of $\mathcal{E}_\theta$ are shared across strokes. We take the final hidden-state feature as the localised representation ${f}^{l}_{s_i} \in \mathbb{R}^{d_s}$ for $i^{th}$ stroke. Thereafter, feature representation of $K$ such strokes having size of $\mathbb{R}^{K \times d_s}$ are further fed to a global relational network ($\mathcal{R}_\theta$) whose final hidden state $f^{g} \in  \mathbb{R}^{d_s}$ captures the global semantic information of the whole sketch. Taking inspiration from residual learning \cite{he2016deep}, we fuse the global feature with individual stroke-level feature through a residual connection with LayerNorm \cite{ba2016layer}. In concrete, every stroke feature enriched by global-local compositional hierarchy is represented by $\hat{f}_{s_i} = \texttt{LayerNorm}({f}^{l}_{s_i} + f^{g}) \in \mathbb{R}^{d}$. We implement both $\mathcal{E}_\theta$ and $\mathcal{R}_\theta$ through a one layer LSTM with hidden state size $128$. Further, we apply a shared linear layer ($\mathcal{C}_\theta$) to get $p(a_i|s_i) = \texttt{softmax} (W_\mathcal{X}\hat{f}_{s_i} + b_{\mathcal{X}})$, where $W_\mathcal{X} \in \mathbb{R}^{d_s\times2}$ and $b_\mathcal{X} \in \mathbb{R}^{2}$. We group three modules  $\{\mathcal{R}_{\theta}, \mathcal{E}_{\theta}, \mathcal{C}_{\theta}\}$ of stroke subset selector as $\mathcal{X_\theta}$. See Fig.~\ref{fig_archi}.

\vspace{-0.1cm}
\subsection{Training Procedure}
\vspace{-0.2cm} 
\keypoint{Necessity of RL:} Due to the unavailability of ground-truth for optimum strokes, we rely on the pre-trained FG-SBIR model to learn the optimum stroke-subset selection strategy.
In particular, given probability distribution $p(a_i|s_i) \in \mathbb{R}^{2}$ for \emph{every} stroke over $\{\texttt{select}, \texttt{ignore}\}$, we can sample from categorical distribution as $a_i \sim  \small{\texttt{Categorical}} \normalsize([p(a_{select}|s_i), p(a_{ignore}|s_i)])$, and thereby we will be getting a stroke subset as $\overline{S}_{V}$ with $K'$ strokes, where $K' \leq K$. In order to get the training signal from pre-trained FG-SBIR model $\mathcal{F}$, we need to feed the subset sketch through $\mathcal{F}$. For that, we need to convert the sequential sketch vector to raster sketch image through rasterization $\overline{S}_{I} = \mathcal{R}(\overline{S}_{V})$, as fine-grained SBIR model only \cite{bhunia2021more, bhunia2020sketch} works on raster image space. While subset sampling could be relaxed by Gumbel-Softmax \cite{jang2016categorical} operation for differentiablity, non-differentiable rasterization operation $\mathcal{R}(\cdot)$ squeeze us to use Policy-Gradient \cite{sutton2000policy} from Reinforcement Learning (RL) literature \cite{kaelbling1996reinforcement}. 

\keypoint{MDP Formulation:}  In particular, given an input sketch $S_{V}$ (initial state), the stroke-subset selector $(\mathcal{X}_{\theta})$ acts a \emph{policy network} which takes \emph{action} on selecting every stroke, and we get an updated state as subset-sketch $\overline{S}_{V}$ (next state). In order to train the policy network, we calculate \emph{reward} using $\mathcal{F}$ as a critic. Therefore, we can form the tuple of four elements \small{(\texttt{initial\_state}, \texttt{action}, \texttt{reward}, \texttt{next\_state})} \normalsize that is typically required to train any RL model. In order to model the existence of multiple possible successful subsets, we unroll this sequential Markov Decision Process (MDP)  $T$ times starting from the complete sketch vector. In other words, for each sketch data, we sequentially sample the subset strokes $T$ times to learn the multi-modal nature of true stroke subsets. Empirically we keep \emph{episode length} $T=5$.

\keypoint{Reward Design:} Our objective is to select the optimum set of stroke which can retrieve the paired photo with minimum rank (e.g. best scenario: rank 1). In other words,  pairwise-distance between the query sketch and paired photo embeddings should be lower than that of query sketch and rest other photos of the gallery.  {As $\mathcal{F}$ is fixed, we can pre-compute the features of all $M$ gallery photos as $G \in \mathbb{R}^{M\times D}$ -- thus eliminating the burden of repeatedly computing the photo features. During stroke subset selector training, we just need to calculate the feature embedding $\mathcal{F}(\overline{S}_{I})$ of rasterized version of predicted subset sketch, and we can calculate rank of paired photo using $G$ and paired-photo index efficiently.} We compute the reward both in the ranking space as well as in the feature embedding space using standard triplet loss on $\mathcal{F}(\overline{S}_{I})$ following Eqn. \ref{eq1}, which is found to give better stability and faster training convergence. In particular, we want to minimise the rank of the paired photo and triplet loss simultaneously. Following the conventional norm of reward maximisation, we define the \emph{reward} (R) as weighted summation of inverse of the rank and negative triplet loss as follows:

\vspace{-0.3cm}
\begin{equation}
\label{eq2}
R = \omega_1 \cdot \frac{1}{rank} + \omega_2 \cdot (-\mathcal{L}_{Triplet})
\end{equation}
\vspace{-0.4cm}

\keypoint{Actor Critic PPO:} We make use of actor-critic version of Proximal Policy Optimisation (PPO) with clipped surrogate objective \cite{schulman2017proximal}  to train our stroke-subset selector. In particular, the very basic policy gradient \cite{sutton2000policy}  objective that is to be minimised could be written as: \cut{$L^{PG}(\theta) = -\frac{1}{K}\sum_{i=1}^{K} \log p_{\theta}(a_i | s_i) \cdot R$ }

\vspace{-0.35cm}
\begin{equation}
\label{eq3}
L^{PG}(\theta) = -\frac{1}{K}\sum_{i=1}^{K} \log p_{\theta}(a_i | s_i) \cdot R
\end{equation}
\vspace{-0.35cm}

For sampling efficiency, using the idea of Importance Sampling \cite{neal2001annealed}, PPO maintains an older policy $p'_{\theta}(a_i | s_i)$, and thus
Conservative Policy Iteration (CPI) objective becomes  $L^{CPI}(\theta) = -\frac{1}{K}\sum_{i=1}^{K} r_i(\theta)\cdot R$, where \cut{$r_i(\theta) =\frac{\log p_{\theta}(a_i | s_i)}{\log p'_{\theta}(a_i | s_i)}$.} $r_i(\theta) = {\log p_{\theta}(a_i | s_i)}/{\log p'_{\theta}(a_i | s_i)}$. Further on, the clipped surrogate objective PPO can be written as $L^{CLIP}(\theta) = -\frac{1}{K}\sum_{i=1}^{K} \texttt{clip} (r_i(\theta), 1-\varepsilon, 1-\varepsilon))$, which aims to penalise too large policy update with hyperparameter $\epsilon=0.2$. We take a minimum of the clipped and unclipped objective, so the final objective is a lower bound (i.e., a pessimistic bound) on the unclipped objective. The final \emph{actor only} version PPO objective becomes:

\vspace{-0.4cm}
\begin{equation}
L^{A}(\theta) = -\frac{1}{K}\sum_{i=1}^{K} \mathrm{min}(L^{CPI}, L^{CLIP})
\vspace{-0.2cm}
\end{equation}

To reduce the variance, the \emph{actor-critic} version of PPO make use of a \emph{learned state-value function} $V(S)$ where $S$ is the sketch vector $S = (s_1, s_2, \cdots, s_K)$. {$V(S)$ shares parameter with actor network $\mathcal{X}_{\theta}$, where only the last linear layer ($C_\theta$) is replaced by a new linear layer upon a single latent vector (accumulated stroke-wise features by averaging), predicting a scalar value that tries to \emph{approximate the reward value}.} Thus, the final loss function combines the policy surrogate and value function error time together with a entropy bonus ($E_n$)  to ensure sufficient exploration is:

\vspace{-0.5cm}
\begin{equation}
L^{AC}(\theta) = -\frac{1}{K}\sum_{i=1}^{K} (L^{A} - c_1 (V_{\theta}(S) - R)^2 + c_2 E_n )
\vspace{-0.2cm}
\end{equation}

where, $c_1$ and $c_2$ are coefficients. As we unroll the sequential stroke-subset selection process for $T=5$, for every sample the loss accumulated over the MDP episode is $\frac{1}{T} \sum_{t=1}^{T}L^{AC}_t(\theta)$.

\vspace{-0.1cm} 
\section{Applications of Stroke-Subset Selector} \label{byproduct}
\vspace{-0.2cm}
 
\keypoint{\emph{Resistance against noisy strokes:}} Collected sketch labels, which are used to train the initial fine-grained SBIR model are also noisy. The proposed stroke-subset selector \emph{not only assists during inference} by noisy-stroke elimination, but also \emph{helps in cleaning training data}, which in turn can boost the performance to some extent. In particular, we train the FG-SBIR model and Stroke-Subset Selector in stage-wise alternative manner, with the FG-SBIR model using clean sketch labels produced by the trained stroke-subset selector. Our method thus offers a plausible way to alleviate the latent/hidden noises of a FG-SBIR dataset \cite{yu2016sketch}.

\keypoint{\emph{Modelling ability to retrieve:}} As the critic network tries to approximate the scalar reward value which is a measure of retrieval performance, we can use the \emph{critic-network} to quantify the retrieval ability at any instant of a sketching episode. Higher scalar score from the critic signifies better retrieval ability. To wit, we ask the question whether a partial sketch is good enough for retrieval or not. Thus, instead of feeding rasterized partial sketch multiple times for on-the-fly \cite{bhunia2020sketch} retrieval, we can save significant computation cost by feeding \emph{only after} it gains a potential retrieval ability. Moreover, as both our actor and critic networks work in sketch vector modality, it adds less computational burden. 


\keypoint{\emph{On-the-fly FG-SBIR: Training from Partial Sketches:}} State-of-the-art on-the-fly FG-SBIR \cite{bhunia2020sketch} employs continuous RL for training using ranking objective. A supervised triplet-loss \cite{yu2017sketch} based training, augmented with synthetic partial sketches obtained through random stroke-dropping is claimed to be sub-optimal, as randomly dropped strokes frequently banish crucial details, resulting in the augmented partial sketch containing insufficient information to depict the paired photo. In contrast, we use our stroke-subset selector to create several augmented partial versions of the same sketch, each with \emph{sufficient retrievability}. While continuous RL is time intensive to train and allegedly unstable \cite{kaelbling1996reinforcement}, we can use simple triplet-loss based supervised learning \cut{to reduce SOTA on-the-fly FG-SBIR performance by creating}with multiple \emph{meaningful augmented partial sketches}.

\vspace{-0.2cm}
\section{Experiments}
\vspace{-0.2cm}
\keypoint{Datasets:} Two publicly available FG-SBIR datasets  \cite{yu2016sketch, pang2019generalising, bhunia2020sketch} namely QMUL-Shoe-V2
and QMUL-Chair-V2 are used in our experiments. Apart from having instance-wise paired sketch-photo, these datasets also contain the sketch coordinate information, and thus would enable us to train the stroke-subset selector using sketch vector modality. We use the standard training/testing split used by the existing state-of-the-arts. In particular, out of $6,730$ ($1,800$) sketches and $2,000$ ($400$) photos from Shoe-V2 (Chair-V2) dataset, $6,051$ sketches ($1,275$) and $1,800$ ($300$) photos are used for training respectively, and the rest are for testing \cite{bhunia2020sketch}.

\keypoint{Implementation:} We have conducted all our experiments on an 11-GB Nvidia RTX 2080-Ti GPU with PyTorch. For \emph{fine-grained SBIR}, we have used  ImageNet \cite{russakovsky2015imagenet} pre-trained VGG-16 \cite{simonyan2014very} backbone with feature embedding dimension $d=512$. We train the FG-SBIR model using Adam optimiser \cite{kingma2014adam} with a learning rate of 0.0001, batch size 16, and margin value of 0.2 for triplet loss. For \emph{stroke subset selector}, we model local stroke embedding network and global relational network using one-layer LSTM with hidden state size $128$ for each. The critic network shares the same weights with that of the actor, with only the last linear layer $C_\theta$ being replaced by a new one that predicts a single scalar value. We train it for $2000$ epoch using Adam optimiser with initial learning rate $10^{-4}$ till $100$ epochs, then reducing to $10^{-5}$. We use a batch size of $16$ and keep an old policy network for importance sampling \cite{neal2001annealed} with episode length $T=5$, and sampled instances are stored in a replay buffer. We update the current policy network at every $20$ iteration using sampled instances from the replay buffer, and the old policy network's weights are copied from the current one for subsequent sampling.  We empirically set both $\omega_1$, $\omega_2$ to $1$, and keep $c_1=0.5$, $c_2=0.01$, $\epsilon=0.2$.


\keypoint{Evaluation Metric:} \textbf{(a) Standard FG-SBIR:} Aligning to the existing state-of-the-art FG-SBIR frameworks \cite{pang2020solving, yu2016sketch}, we use percentage of sketches having true-matched photo in the top-1 (acc.@1) and top-5 (acc.@5) lists to assess the FG-SBIR performance. \textbf{(b) On-the-fly FG-SBIR:} Furthermore, to showcase the early retrieval performance from partial sketch, adhering to prior early-retrieval work \cite{bhunia2020sketch} we employ two plots namely, (i)\emph{ranking-percentile}  and (ii)  $\frac{1}{rank}$ vs. \emph{percentage of sketch}. Higher area under these curves indicate better early-retrieval potential. For the sake of simplicity, we call area under curves (i) and (ii) as r@A and r@B through the rest of the paper.

\keypoint{Competitors:} To the best of our knowledge, no earlier works have directly attempted to design a Noise-Tolerant FG-SBIR model in the SBIR literature.  Therefore, we compare with the existing standard FG-SBIR works appeared in the literature, as well as, we develop some self-designed competitive baselines under the assumption of \emph{`all sketches are sketchy'} -- which explicitly intend to learn invariance against noisy strokes. {(a)}  \emph{State-of-the-arts} (SOTA): While \textbf{Triplet-SN} \cite{yu2016sketch} uses Sketch-A-Net backbone along with triplet loss,  \textbf{Triplet-Attn-HOLEF}  extends \cite{yu2016sketch} with spatial attention and higher order ranking loss. Recent works include: \textbf{Jigsaw-Pretrain} with self-supervised pre-training, \textbf{Triplet-RL} \cite{bhunia2020sketch}  employing RL-based fine-tuning, \textbf{StyleMeUP} involving MAML training,  \textbf{Semi-Sup} \cite{bhunia2021more} incorporating semi-supervised paradigm, and \textbf{Cross-Hier} \cite{sain2020cross}  utilising cross-modal hierarchy with costly paired-embedding. {(b)}  \emph{Self-designed Baselines (BL)}:  We  create multiple version of the same sketch by randomly dropping strokes (ensuring percentage of sketch vector length never drops below 80\%) or  by synthetically adding random noisy stroke patches similar to \cite{liu2020unsupervised}. \textbf{Augment} aims to learn the invariance against noisy stroke by adding them inside training. This is further advanced by \textbf{StyleMeUp+Augment} where synthetic noisy/augmented sketches are mixed in the inner-loop of \cite{sain2021stylemeup} to learn invariance by optimising outer-loop on real sketches. \textbf{Contrastive+Augment} imposes an additional contrastive loss \cite{chen2020simple} such that the distance between two augmented versions of same sketch should be lower than that of with a random other sketch. Our pre-trained baseline FG-SBIR model is termed as \textbf{B-Siamese}.

\setlength{\tabcolsep}{3pt}
\begin{table}[]
    \centering
    \caption{Results under Standard FG-SBIR setup.}
    \vspace{-0.3cm}
    \footnotesize
    \begin{tabular}{ccccccc}
        \hline
         &   &   & \multicolumn{2}{c}{Chair-V2} & \multicolumn{2}{c}{Shoe-V2} \\
        \cline{4-7}
         & & & Acc@1 & Acc@5 & Acc@1 & Acc@5\\
        \hline
        \multirow{7}{*}{\rotatebox[origin=c]{0}{SOTA}} 
         & \multicolumn{2}{c}{Triplet-SN \cite{yu2016sketch}} & 47.4\% & 71.4\% & 28.7\% & 63.5\% \\
         & \multicolumn{2}{c}{Triplet-Attn-HOLEF \cite{song2017deep}} & 50.7\% & 73.6\% & 31.2\% & 66.6\% \\
        & \multicolumn{2}{c}{Triplet-RL  \cite{bhunia2020sketch}} & 51.2\% & 73.8\% & 30.8\% & 65.1\% \\ 
         & \multicolumn{2}{c}{Mixed-Jigsaw \cite{pang2019generalising}} & 56.1\% & 75.3\% & 36.5\% & 68.9\% \\
          & \multicolumn{2}{c}{Semi-Sup \cite{bhunia2021more}} & 60.2\% & 78.1\% & 39.1\% & 69.9\% \\
          & \multicolumn{2}{c}{StyleMeUp \cite{sain2021stylemeup}} & 62.8\% & 79.6\% & 36.4\% & 68.1\% \\
           & \multicolumn{2}{c}{Cross-Hier \cite{sain2020cross}} & {62.4\%} & {79.1\%} & {36.2\%} & {67.8\%} \\

         \hdashline
        \multirow{4}{*}{\rotatebox[origin=c]{0}{BL}} 
        & \multicolumn{2}{c}{\blue{(B)aseline-Siamese}} & {\blue{53.3\%}} & {\blue{74.3\%}} & {\blue{33.4\%}} & {\blue{67.8\%}}\\
         & \multicolumn{2}{c}{Augmnt} & {54.1\%} & {74.6\%} & {33.9\%} & {68.2\%}\\
          & \multicolumn{2}{c}{StyleMeUp+Augment} & {56.1\%} & {76.9\%} & {36.9\%} & {69.9\%}\\
        & \multicolumn{2}{c}{Contrastive+Augment} & {58.8\%} & {77.1\%} & {37.6\%} & {70.1\%}\\
        \hdashline
        \multirow{2}{*}{\rotatebox[origin=c]{0}{Limits}} 
        & \multicolumn{2}{c}{Upper-Limit} & 78.6\% & 90.3\% & 66.3\% & 88.3\% \\
         & \multicolumn{2}{c}{Linear-Limit} & 59.4\% & 77.3\% & 42.5\% & 73.2\% \\
         \hdashline
        & \multicolumn{2}{c}{Proposed} & \red{64.8\%} & \red{79.1\%} & \red{43.7\%} & \red{74.9\%} \\
        \hline
    \end{tabular}
    \label{tab:my_label2}
        \vspace{-0.5cm}
\end{table}

\vspace{-0.1cm}
\subsection{Performance Analysis}
\vspace{-0.1cm}
The comparative analysis is shown in Table \ref{tab:my_label2}. Overall, we observe a significantly improved performance of our proposed Noise-Resistant fine-grained SBIR employing a stroke-subset selector as a pre-processing neural agent compared to the existing state-of-the-art. The early works tried to address different \emph{architectural modifications} \cite{song2017fine, pang2019generalising}, and later on the field of fine-grained SBIR witnessed successive improvements through adaptation of different paradigms like \emph{self-supervised learning} \cite{pang2020solving}, \emph{meta-learning} \cite{sain2021stylemeup}, \emph{semi-supervised learning} \cite{bhunia2021more}, etc. As opposed to these works, we underpin an important phenomenon of noisy strokes, which is inherent to FG-SBIR. Most interestingly, our simple stroke-subset selector can improve the performance of baseline B-Siamese model by an approximate margin of $10.31\%$ without any complicated joint-training of \emph{Semi-Sup} \cite{bhunia2021more}, costly hierarchical paired embedding of \emph{Cross-Hier} \cite{sain2020cross}, or meta-learning cumbersome feature transformation layer of \emph{StyleMeUp} \cite{sain2021stylemeup}. Furthermore, the performance of \emph{Augmnt} baseline is slightly better than our baseline pre-trained FG-SBIR as it learns some invariance from augmented/partial sketch. While we experienced difficulty in stable training for \emph{StyleMeUp+Augment}, \emph{Contrastive+Augment} appears as a simple and straightforward way to learn the invariance against noisy strokes. Instead of modelling invariance, we aim to eliminate the noisy strokes, thus giving a freedom of explainability through visualisation. Despite using complicated architectures \cite{sain2020cross, bhunia2021more}, SOTA fails even to beat the accuracy of Linear-Limit (refer to section \ref{sec3}), while we can. Nevertheless, we suppress it by keeping the simple baseline FG-SBIR untouched and prepending a simple stroke-selector agent -- working on a cheaper vector modality for efficient deployment.

\vspace{-0.1cm}
\subsection{Further Analysis and Insights}
\label{ablation}
\vspace{-0.1cm}
\noindent \textbf{Ability to retrieve/classify for partial sketches:} 
\cut{As described in Section \ref{byproduct}, we first aim to validate the potential of auxiliary critic network to quantify the retrieval ability of partial sketches.} The scalar value predicted by our learned state-value function (critic-network) \cite{schulman2017proximal} signifies the retrieval ability of partial sketch with the notion of higher being the better. We here train our model with a reward of $\frac{1}{rank}$ for easy interpretability. Once the stroke-subset selector with actor-critic version is trained, we feed the sketch to the critic network (in vector space) at a progressive step of $5\%$ completion, and record the predicted scalar value at every instant. At the same time, we rasterize every partial instance and feed through pre-trained FG-SBIR to calculate the resultant ranking percentile of the paired photo. In Fig.~\ref{fig_partial}, the high correlation demonstrates that the partial sketch with a higher scalar score by the critic network tends to have a higher average ranking percentile (ARP), while those with a lesser score result in lower ARP. Quantitatively, the top@5 accuracy for partial sketches is $80.1\%$, which have a higher predicted scalar score than a threshold of $\frac{1}{5}$. This validates the potential of our critic network in quantifying if a partial sketch is sufficient for retrieval. Suppose we repeat the same with the negative of the classification loss as a reward for a pre-trained classification network. In that case as well, we observe a similar consistent behaviour for partial sketch classification, indicating our approach to be generic for various sketch-related downstream tasks. See \S~supplementary.

\vspace{-0.2cm}
\begin{figure}[!hbt]
\centering
\includegraphics[width=\columnwidth,height=3cm]{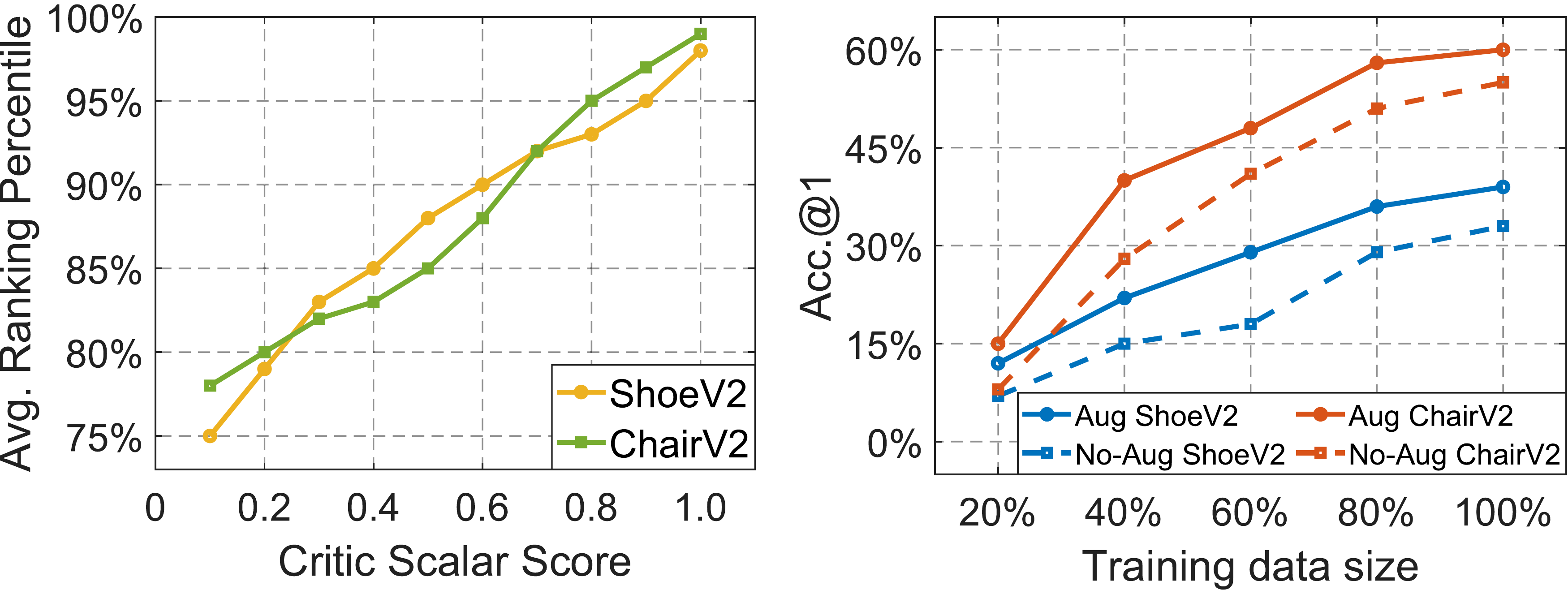}
\vspace{-0.8cm}
\caption{ (a) Retrieval ability of partial sketch: correlation between critic network $\mathrm{V(S)}$ predicted score and ranking percentile (b) Performance at varying training data size with stroke-subset selector based data augmentation.}
\label{fig_partial}
\vspace{-0.4cm}
\end{figure}



\keypoint{Data Augmentation:} Our elementary study reveals that there exists multiple possible subsets which can retrieve the paired photo faithfully. In particular, we use our policy network to get stroke wise importance measure using $p(a_i|s_i)$ towards the retrieval objectives. Through categorical sampling of $p(a_i|s_i)$, we can create multiple augmented versions of the same sketch to increase the training data size. To validate this, we compute the performance of baseline retrieval model at varying training data size with our sketch augmentation strategy in Fig.~\ref{fig_partial}. While accuracy remains marginally better towards the high data regime, stroke subset selection based strategy excels the standard supervised counter-part by a significant margin, thus proving the efficacy of our smart data-augmentation approach.  





\begin{figure}[!hbt]
\includegraphics[width=\columnwidth,height=3.2cm]{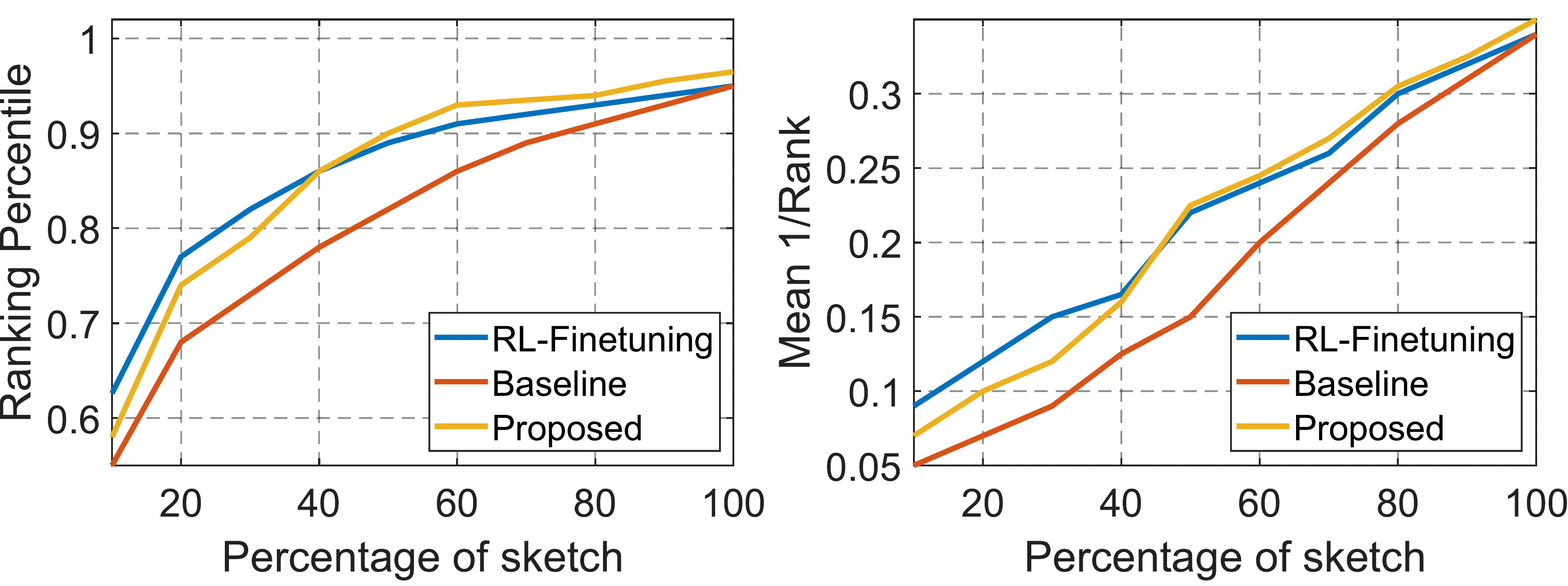}
\vspace{-0.75cm}
\caption{Comparative results under on-the-fly setup (Shoe-V2), visualised through percentage of sketch. Higher area under the plots indicates better early retrieval performance.}
\label{fig_onthefly}
\vspace{-0.4cm}
\end{figure}

\keypoint{On-the-Fly Retrieval:}  \cut{Training a model partial sketches with random stroke-dropping gives rise to noisy gradient, and thus this naive baseline falls short compared to RL-based fine-tuning that consider the complete sketch drawing episode for training.} {Training a model with partial sketches generated by \emph{random stroke-dropping} gives rise to noisy gradient, and thus this naive baseline falls short compared to RL-based fine-tuning that consider the complete sketch drawing episode for training.} In lieu of RL-based fine-tuning \cite{bhunia2020sketch}, we train an on-the-fly retrieval model from meaningful (holds ability to retrieve) partial sketches augmented through our critic network that have a higher scalar score than $\frac{1}{20}$. While training a continuous RL pipeline  \cite{bhunia2020sketch} is unstable and time-consuming, we achieve a competitive on-the-fly \emph{r@A(r@B)} performance  of $85.78$($21.1$) with \emph{basic} triplet-loss based model trained with \emph{smartly} augmented partial sketches compared to $85.38$ ($21.24$) as claimed in \cite{bhunia2020sketch} on ShoeV2. From Fig.~\ref{fig_onthefly}, we can see that at very early few instances, RL-Based fine-tuning \cite{bhunia2020sketch} performs better, while ours achieve a significantly better performance as the drawing episode proceeds towards completion. While early sketch drawing episode is too coarse that hardly it can retrieve, through modelling the retrieval ability (with threshold of $\frac{1}{10}$) of partial sketches, we can reduce the number of time we need to feed the rasterized sketch by $42.2\%$ with very little drop in performance (r@A(r@B): $85.07$ ($20.98$)). Thus modelling partial sketches lead to significant computational edge under on-the-fly setting.  




\vspace{-0.3cm}

\begin{figure}[!hbt]
\centering
\includegraphics[height=2.5cm, width=\linewidth]{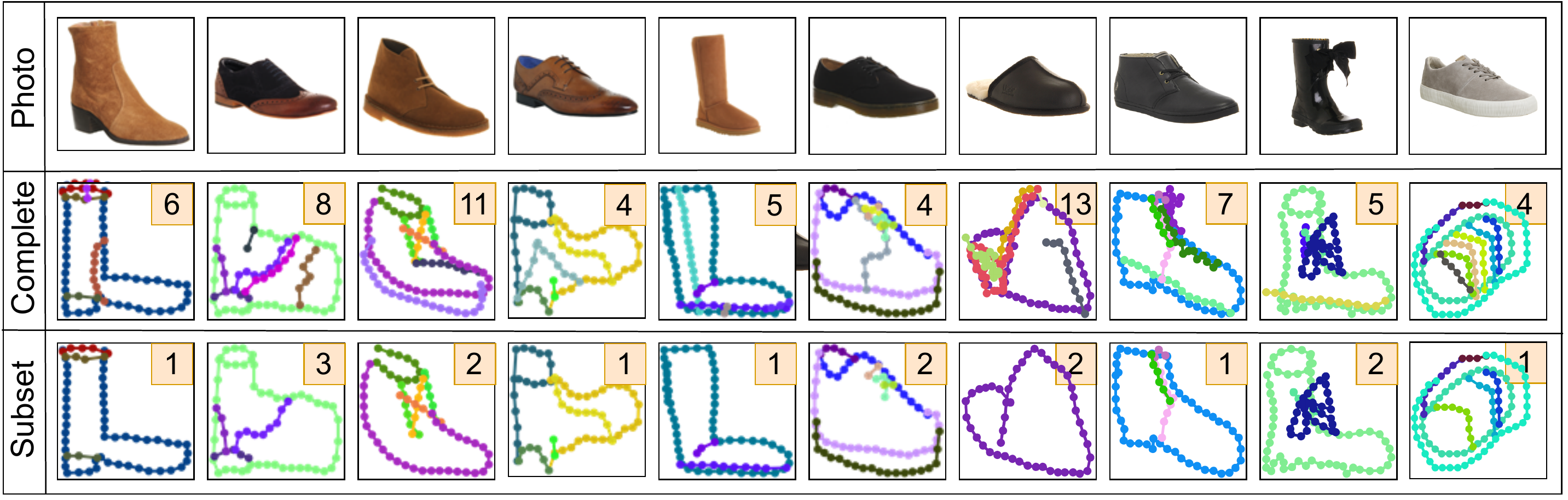}
\vspace{-0.8cm}
\caption{Examples showing selected subset performing better (rank in box) than complete sketch from ShoeV2.}
\label{fig6}
\vspace{-0.3cm}
\end{figure}

\begin{figure}[!hbt]
\centering
\includegraphics[height=2.5cm, width=\linewidth]{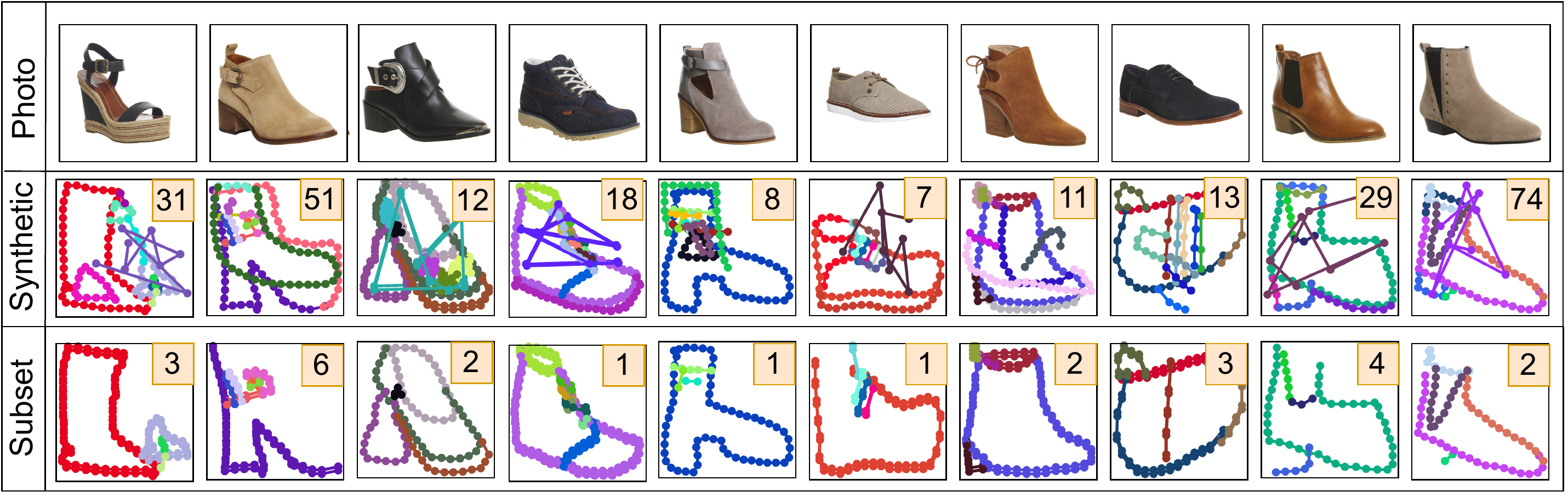}
\vspace{-0.8cm}
\caption{Examples showing ability to perform (rank in box) under \emph{synthetic} noisy sketch input on ShoeV2.}
\label{fig7}
\vspace{-0.6cm}
\end{figure}

\keypoint{Resistance to Noisy Stroke:} The significance of stroke subset selector is quantitatively shown in Table~\ref{tab:my_label2}.  While it validates our potential under inherent low-magnitude noise existed in the dataset (shown in Fig.~\ref{fig6}), we further aim to see how our method works on extreme noisy situation.  In particular, we augment the training sketches by synthetic noisy patches, and train our subset selector with a pre-trained retrieval model. During testing, we synthetically add noisy strokes \cite{liu2020unsupervised}, and pass it through stroke-subset selector (pre-processing module) before feeding it to the retrieval model. While excluding the selector, the top@1 (top@5) drops to $13.4\%$($44.9\%$) in presence of synthetic noises, our stroke subset selector can improve them to $37.2\%$($68.2\%$) by eliminating the synthetic noisy strokes (see Fig.~\ref{fig7}).



\keypoint{Ablation on Design:} (i)  Instead of designing the stroke subset selector through hierarchical LSTM, another straight forward way is to use one layer bidirectional LSTM, where every coordinate point is being fed to each time step. However, the top@1(top@5) lags behind by $4.9\%$($6.7\%$) than ours, which verifies the necessity of hierarchical modelling of sketch vectors to consider the compositional relationship in our problem. Replacing LSTM by Transformer leads to no meaningful improvement in our case.  (ii) Being a pre-processing step, we compare the extra time required for selecting the optimal stroke set. In particular, it adds extra \doublecheck{$22.4\%$} multiply-add operations and $18.3\%$ extra CPU time compared standard baseline FG-SBIR. 
(iii) Compared to different RL methods \cite{schulman2017proximal},  we get best results with PPO actor-critic version with clipped surrogate objective that beats its actor-only alternative by $1.7\%$ top@1 accuracy(ShoeV2). Importantly, training with critic network leads to one important byproduct of modelling retrieval ability of partial sketches. 
(iv) Exploring different possible reward functions, we conclude that combining rewards from both ranking and feature embedding space through triplet loss gives most optimum performance than ranking only counterpart by extra $1.2\%$ top@1 accuracy (ShoeV2). Please refer to supplementary for more details. 





\vspace{-0.2cm} 
\section{Conclusion}
\vspace{-0.1cm}
In this paper, we tackle the ``fear to sketch'' issue by proposing an intelligent stroke subset selector that automatically selects the most representative stroke subset from the entire query stroke set. Our stroke subset selector can detect and eliminate irrelevant (noisy) strokes, thus boosting performance of any off-the-shelf FG-SBIR framework. To this end, we designed an RL-based framework, which learns to form an optimal stroke subset by interacting with a pre-trained FG-SBIR model. We also show how the proposed selector can augment other sketch applications in a plug-and-play manner.


{\small
\bibliographystyle{ieee_fullname}
\bibliography{main}

\begin{thebibliography}{10}\itemsep=-1pt

\bibitem{aksan2020cose}
Emre Aksan, Thomas Deselaers, Andrea Tagliasacchi, and Otmar Hilliges.
\newblock Cose: Compositional stroke embeddings.
\newblock In {\em NeurIPS}, 2021.

\bibitem{arulkumaran2017deep}
Kai Arulkumaran, Marc~Peter Deisenroth, Miles Brundage, and Anil~Anthony
  Bharath.
\newblock Deep reinforcement learning: A brief survey.
\newblock {\em IEEE Signal Processing Magazine}, 2017.

\bibitem{ba2016layer}
Jimmy~Lei Ba, Jamie~Ryan Kiros, and Geoffrey~E Hinton.
\newblock Layer normalization.
\newblock {\em arXiv preprint arXiv:1607.06450}, 2016.

\bibitem{bhunia2021more}
Ayan~Kumar Bhunia, Pinaki~Nath Chowdhury, Aneeshan Sain, Yongxin Yang, Tao
  Xiang, and Yi-Zhe Song.
\newblock More photos are all you need: Semi-supervised learning for
  fine-grained sketch based image retrieval.
\newblock In {\em CVPR}, 2021.

\bibitem{bhunia2021vectorization}
Ayan~Kumar Bhunia, Pinaki~Nath Chowdhury, Yongxin Yang, Timothy~M Hospedales,
  Tao Xiang, and Yi-Zhe Song.
\newblock Vectorization and rasterization: Self-supervised learning for sketch
  and handwriting.
\newblock In {\em CVPR}, 2021.

\bibitem{bhunia2020pixelor}
Ayan~Kumar Bhunia, Ayan Das, Umar~Riaz Muhammad, Yongxin Yang, Timothy~M
  Hospedales, Tao Xiang, Yulia Gryaditskaya, and Yi-Zhe Song.
\newblock Pixelor: A competitive sketching ai agent. so you think you can
  sketch?
\newblock {\em ACM-TOG}, 2020.

\bibitem{DoodleIncremental}
Ayan~Kumar Bhunia, Viswanatha~Reddy Gajjala, Subhadeep Koley, Rohit Kundu,
  Aneeshan Sain, Tao Xiang, and Yi-Zhe Song.
\newblock Doodle it yourself: Class incremental learning by drawing a few
  sketches.
\newblock In {\em CVPR}, 2022.

\bibitem{bhunia2020sketch}
Ayan~Kumar Bhunia, Yongxin Yang, Timothy~M Hospedales, Tao Xiang, and Yi-Zhe
  Song.
\newblock Sketch less for more: On-the-fly fine-grained sketch-based image
  retrieval.
\newblock In {\em CVPR}, 2020.

\bibitem{chen2020simple}
Ting Chen, Simon Kornblith, Mohammad Norouzi, and Geoffrey Hinton.
\newblock A simple framework for contrastive learning of visual
  representations.
\newblock In {\em ICML}, 2020.

\bibitem{chen2018sketchygan}
Wengling Chen and James Hays.
\newblock Sketchygan: Towards diverse and realistic sketch to image synthesis.
\newblock In {\em CVPR}, 2018.

\bibitem{PartialSBIR}
Pinaki~Nath Chowdhury, Ayan~Kumar Bhunia, Viswanatha~Reddy Gajjala, Aneeshan
  Sain, Tao Xiang, and Yi-Zhe Song.
\newblock Partially does it: Towards scene-level fg-sbir with partial input.
\newblock In {\em CVPR}, 2022.

\bibitem{collomosse2019livesketch}
John Collomosse, Tu Bui, and Hailin Jin.
\newblock Livesketch: Query perturbations for guided sketch-based visual
  search.
\newblock In {\em CVPR}, 2019.

\bibitem{dey2019doodle}
Sounak Dey, Pau Riba, Anjan Dutta, Josep Llados, and Yi-Zhe Song.
\newblock Doodle to search: Practical zero-shot sketch-based image retrieval.
\newblock In {\em CVPR}, 2019.

\bibitem{duong2019automatic}
Chi~Nhan Duong, Khoa Luu, Kha~Gia Quach, Nghia Nguyen, Eric Patterson, Tien~D
  Bui, and Ngan Le.
\newblock Automatic face aging in videos via deep reinforcement learning.
\newblock In {\em CVPR}, 2019.

\bibitem{dutta2019semantically}
Anjan Dutta and Zeynep Akata.
\newblock Semantically tied paired cycle consistency for zero-shot sketch-based
  image retrieval.
\newblock In {\em CVPR}, 2019.

\bibitem{ghosh2019interactive}
Arnab Ghosh, Richard Zhang, Puneet~K Dokania, Oliver Wang, Alexei~A Efros,
  Philip~HS Torr, and Eli Shechtman.
\newblock Interactive sketch \& fill: Multiclass sketch-to-image translation.
\newblock In {\em ICCV}, 2019.

\bibitem{ha2017neural}
David Ha and Douglas Eck.
\newblock A neural representation of sketch drawings.
\newblock In {\em ICLR}, 2018.

\bibitem{han2018masking}
Bo Han, Jiangchao Yao, Gang Niu, Mingyuan Zhou, Ivor Tsang, Ya Zhang, and
  Masashi Sugiyama.
\newblock Masking: A new perspective of noisy supervision.
\newblock In {\em NeurIPS}, 2018.

\bibitem{han2019deep}
Xiaoguang Han, Zhaoxuan Zhang, Dong Du, Mingdai Yang, Jingming Yu, Pan Pan, Xin
  Yang, Ligang Liu, Zixiang Xiong, and Shuguang Cui.
\newblock Deep reinforcement learning of volume-guided progressive view
  inpainting for 3d point scene completion from a single depth image.
\newblock In {\em CVPR}, 2019.

\bibitem{he2016deep}
Kaiming He, Xiangyu Zhang, Shaoqing Ren, and Jian Sun.
\newblock Deep residual learning for image recognition.
\newblock In {\em CVPR}, 2016.

\bibitem{jang2016categorical}
Eric Jang, Shixiang Gu, and Ben Poole.
\newblock Categorical reparameterization with gumbel-softmax.
\newblock In {\em ICLR}, 2017.

\bibitem{jo2019sc}
Youngjoo Jo and Jongyoul Park.
\newblock Sc-fegan: Face editing generative adversarial network with user's
  sketch and color.
\newblock In {\em ICCV}, 2019.

\bibitem{kaelbling1996reinforcement}
Leslie~Pack Kaelbling, Michael~L Littman, and Andrew~W Moore.
\newblock Reinforcement learning: A survey.
\newblock {\em JAIR}, 1996.

\bibitem{simonyan2015very}
Andrew~Zisserman Karen~Simonyan.
\newblock Very deep convolutional networks for large-scale image recognition.
\newblock In {\em ICLR}, 2015.

\bibitem{kingma2014adam}
Diederik~P Kingma and Jimmy Ba.
\newblock Adam: A method for stochastic optimization.
\newblock In {\em ICLR}, 2014.

\bibitem{lee2018stacked}
Kuang-Huei Lee, Xi Chen, Gang Hua, Houdong Hu, and Xiaodong He.
\newblock Stacked cross attention for image-text matching.
\newblock In {\em ECCV}, 2018.

\bibitem{liang2017deep}
Xiaodan Liang, Lisa Lee, and Eric~P Xing.
\newblock Deep variation-structured reinforcement learning for visual
  relationship and attribute detection.
\newblock In {\em CVPR}, 2017.

\bibitem{liu2019sketchgan}
Fang Liu, Xiaoming Deng, Yu-Kun Lai, Yong-Jin Liu, Cuixia Ma, and Hongan Wang.
\newblock Sketchgan: Joint sketch completion and recognition with generative
  adversarial network.
\newblock In {\em CVPR}, 2019.

\bibitem{liu2017deep}
Li Liu, Fumin Shen, Yuming Shen, Xianglong Liu, and Ling Shao.
\newblock Deep sketch hashing: Fast free-hand sketch-based image retrieval.
\newblock In {\em CVPR}, 2017.

\bibitem{liu2020unsupervised}
Runtao Liu, Qian Yu, and Stella~X Yu.
\newblock Unsupervised sketch to photo synthesis.
\newblock In {\em ECCV}, 2020.

\bibitem{muhammad2019goal}
Umar~Riaz Muhammad, Yongxin Yang, Timothy~M Hospedales, Tao Xiang, and Yi-Zhe
  Song.
\newblock Goal-driven sequential data abstraction.
\newblock In {\em ICCV}, 2019.

\bibitem{muhammad2018learning}
Umar~Riaz Muhammad, Yongxin Yang, Yi-Zhe Song, Tao Xiang, and Timothy~M
  Hospedales.
\newblock Learning deep sketch abstraction.
\newblock In {\em CVPR}, 2018.

\bibitem{neal2001annealed}
Radford~M Neal.
\newblock Annealed importance sampling.
\newblock {\em Statistics and Computing}, 2001.

\bibitem{pang2019generalising}
Kaiyue Pang, Ke Li, Yongxin Yang, Honggang Zhang, Timothy~M Hospedales, Tao
  Xiang, and Yi-Zhe Song.
\newblock Generalising fine-grained sketch-based image retrieval.
\newblock In {\em CVPR}, 2019.

\bibitem{pang2017cross}
Kaiyue Pang, Yi-Zhe Song, Tony Xiang, and Timothy~M Hospedales.
\newblock Cross-domain generative learning for fine-grained sketch-based image
  retrieval.
\newblock In {\em BMVC}, 2017.

\bibitem{pang2020solving}
Kaiyue Pang, Yongxin Yang, Timothy~M Hospedales, Tao Xiang, and Yi-Zhe Song.
\newblock Solving mixed-modal jigsaw puzzle for fine-grained sketch-based image
  retrieval.
\newblock In {\em CVPR}, 2020.

\bibitem{ribeiro2020sketchformer}
Leo Sampaio~Ferraz Ribeiro, Tu Bui, John Collomosse, and Moacir Ponti.
\newblock Sketchformer: Transformer-based representation for sketched
  structure.
\newblock In {\em CVPR}, 2020.

\bibitem{russakovsky2015imagenet}
Olga Russakovsky, Jia Deng, Hao Su, Jonathan Krause, Sanjeev Satheesh, Sean Ma,
  Zhiheng Huang, Andrej Karpathy, Aditya Khosla, Michael Bernstein, et~al.
\newblock Imagenet large scale visual recognition challenge.
\newblock {\em IJCV}, 2015.

\bibitem{Sketch3T}
Aneeshan Sain, Ayan~Kumar Bhunia, Vaishnav Potlapalli, Pinaki~Nath Chowdhury,
  Tao Xiang, and Yi-Zhe Song.
\newblock Sketch3t: Test-time training for zero-shot sbir.
\newblock In {\em CVPR}, 2022.

\bibitem{sain2020cross}
Aneeshan Sain, Ayan~Kumar Bhunia, Yongxin Yang, Tao Xiang, and Yi-Zhe Song.
\newblock Cross-modal hierarchical modelling for fine-grained sketch based
  image retrieval.
\newblock In {\em BMVC}, 2020.

\bibitem{sain2021stylemeup}
Aneeshan Sain, Ayan~Kumar Bhunia, Yongxin Yang, Tao Xiang, and Yi-Zhe Song.
\newblock Stylemeup: Towards style-agnostic sketch-based image retrieval.
\newblock In {\em CVPR}, 2021.

\bibitem{schulman2017proximal}
John Schulman, Filip Wolski, Prafulla Dhariwal, Alec Radford, and Oleg Klimov.
\newblock Proximal policy optimization algorithms.
\newblock {\em arXiv preprint arXiv:1707.06347}, 2017.

\bibitem{shen2018zero}
Yuming Shen, Li Liu, Fumin Shen, and Ling Shao.
\newblock Zero-shot sketch-image hashing.
\newblock In {\em CVPR}, 2018.

\bibitem{simonyan2014very}
Karen Simonyan and Andrew Zisserman.
\newblock Very deep convolutional networks for large-scale image recognition.
\newblock {\em arXiv preprint arXiv:1409.1556}, 2014.

\bibitem{song2021learning}
Hwanjun Song, Minseok Kim, Dongmin Park, Yooju Shin, and Jae-Gil Lee.
\newblock Learning from noisy labels with deep neural networks: A survey.
\newblock {\em arXiv preprint arXiv:2007.08199}, 2021.

\bibitem{song2017fine}
Jifei Song, Yi-Zhe Song, Tony Xiang, and Timothy~M Hospedales.
\newblock Fine-grained image retrieval: the text/sketch input dilemma.
\newblock In {\em BMVC}, 2017.

\bibitem{song2017deep}
Jifei Song, Qian Yu, Yi-Zhe Song, Tao Xiang, and Timothy~M Hospedales.
\newblock Deep spatial-semantic attention for fine-grained sketch-based image
  retrieval.
\newblock In {\em CVPR}, 2017.

\bibitem{sutton2000policy}
Richard~S Sutton, David~A McAllester, Satinder~P Singh, and Yishay Mansour.
\newblock Policy gradient methods for reinforcement learning with function
  approximation.
\newblock In {\em NeurIPS}, 2000.

\bibitem{tanno20219learning}
Ryutaro Tanno, Ardavan Saeedi, Swami Sankaranarayanan, Daniel~C Alexander, and
  Nathan Silberman.
\newblock Learning from noisy labels by regularized estimation of annotator
  confusion.
\newblock In {\em CVPR}, 2019.

\bibitem{tolias2017asymmetric}
Giorgos Tolias and Ondrej Chum.
\newblock Asymmetric feature maps with application to sketch based retrieval.
\newblock In {\em CVPR}, 2017.

\bibitem{wang2021sketch}
Sheng-Yu Wang, David Bau, and Jun-Yan Zhu.
\newblock Sketch your own gan.
\newblock In {\em ICCV}, 2021.

\bibitem{wang2019reinforced}
Xin Wang, Qiuyuan Huang, Asli Celikyilmaz, Jianfeng Gao, Dinghan Shen,
  Yuan-Fang Wang, William~Yang Wang, and Lei Zhang.
\newblock Reinforced cross-modal matching and self-supervised imitation
  learning for vision-language navigation.
\newblock In {\em CVPR}, 2019.

\bibitem{weinberger2009distance}
Kilian~Q Weinberger and Lawrence~K Saul.
\newblock Distance metric learning for large margin nearest neighbor
  classification.
\newblock {\em JMLR}, 2009.

\bibitem{xu2018sketchmate}
Peng Xu, Yongye Huang, Tongtong Yuan, Kaiyue Pang, Yi-Zhe Song, Tao Xiang,
  Timothy~M Hospedales, Zhanyu Ma, and Jun Guo.
\newblock Sketchmate: Deep hashing for million-scale human sketch retrieval.
\newblock In {\em CVPR}, 2018.

\bibitem{xu2021multigraph}
Peng Xu, Chaitanya~K Joshi, and Xavier Bresson.
\newblock Multigraph transformer for free-hand sketch recognition.
\newblock {\em IEEE T-NNLS}, 2021.

\bibitem{yelamarthi2018zero}
Sasi~Kiran Yelamarthi, Shiva~Krishna Reddy, Ashish Mishra, and Anurag Mittal.
\newblock A zero-shot framework for sketch based image retrieval.
\newblock In {\em ECCV}, 2018.

\bibitem{yu2016sketch}
Qian Yu, Feng Liu, Yi-Zhe Song, Tao Xiang, Timothy~M Hospedales, and
  Chen-Change Loy.
\newblock Sketch me that shoe.
\newblock In {\em CVPR}, 2016.

\bibitem{yu2021fine}
Qian Yu, Jifei Song, Yi-Zhe Song, Tao Xiang, and Timothy~M Hospedales.
\newblock Fine-grained instance-level sketch-based image retrieval.
\newblock {\em IJCV}, 2021.

\bibitem{yu2017sketch}
Qian Yu, Yongxin Yang, Feng Liu, Yi-Zhe Song, Tao Xiang, and Timothy~M
  Hospedales.
\newblock Sketch-a-net: A deep neural network that beats humans.
\newblock {\em IJCV}, 2017.

\bibitem{zhang2018generative}
Jingyi Zhang, Fumin Shen, Li Liu, Fan Zhu, Mengyang Yu, Ling Shao, Heng~Tao
  Shen, and Luc~Van Gool.
\newblock Generative domain-migration hashing for sketch-to-image retrieval.
\newblock In {\em ECCV}, 2018.

\bibitem{zhou2021robust}
Tianyi Zhou, Shengjie Wang, and Jeff Bilmes.
\newblock Robust curriculum learning: from clean label detection to noisy label
  self-correction.
\newblock In {\em ICLR}, 2021.

\end{thebibliography}
}
\onecolumn{
\begin{center}
\title{\Large{\textbf{Supplementary material for \\ Sketching \emph{without} Worrying: Noise-Tolerant Sketch-Based \\ Image Retrieval}}}

\author{Ayan Kumar Bhunia\textsuperscript{1} \hspace{.2cm}  \hspace{.1cm}  Subhadeep Koley\textsuperscript{1,2} \hspace{.1cm}  Abdullah Faiz Ur Rahman Khilji\thanks{Interned with SketchX} \\ Aneeshan Sain\textsuperscript{1,2} \hspace{.1cm}   Pinaki nath Chowdhury \textsuperscript{1,2} \hspace{.1cm} 
Tao Xiang\textsuperscript{1,2}\hspace{.2cm}  Yi-Zhe Song\textsuperscript{1,2} \\
\textsuperscript{1}SketchX, CVSSP, University of Surrey, United Kingdom.  \\
\textsuperscript{2}iFlyTek-Surrey Joint Research Centre on Artificial Intelligence.\\
{\tt\small \{a.bhunia, s.koley, a.sain, p.chowdhury, t.xiang, y.song\}@surrey.ac.uk} 
}
\end{center}
}

\maketitle


\section{Comparative Study with different RL methods} 

We compare with different RL methods \cite{schulman2017proximal, arulkumaran2017deep}, starting from Vanilla Policy Gradient, Deep Q-Learning, TRPO, to variants of PPO. For our use-case we get best results (Table~\ref{tab:my_label1}) with PPO actor-critic version with clipped surrogate objective, where the critic network leads to one important byproduct of modelling retrieval ability of partial sketches. 

\setlength{\tabcolsep}{4.5pt}
\begin{table}[!hbt]
    \centering
    \caption{Performance analysis using different RL approaches.}
    \footnotesize
    \begin{tabular}{ccccccc}
        \hline
         &   &   & \multicolumn{2}{c}{Chair-V2} & \multicolumn{2}{c}{Shoe-V2} \\
        \cline{4-7}
         & & & Acc@1 & Acc@5 & Acc@1 & Acc@5\\
        \hline
         & \multicolumn{2}{c}{Vanilla Policy Gradient} & 61.4 \% & 78.6\% & 40.1\% & 71.9\% \\
         & \multicolumn{2}{c}{Deep Q-Learning} & 61.9\% & 78.9\% & 40.7\% & 71.8\% \\
        & \multicolumn{2}{c}{TRPO} & 60.8\% & 78.2\% & 39.8\% & 70.2\% \\ 
         & \multicolumn{2}{c}{PPO Actor-Only KL} & 62.8\% & 78.5\% & 42.3\% & 72.8\% \\
          & \multicolumn{2}{c}{PPO Actor-Only Clipped} & 63.9\% & 78.9\% & 43.1\% & 74.5\% \\
          & \multicolumn{2}{c}{PPO Actor-Critic KL} & 63.8\% & 78.7\% & 42.1\% & 73.7\% \\
          & \multicolumn{2}{c}{PPO Actor-Critic Clipped} & 64.8\% & 79.1\% & 43.7\% & 74.9\% \\
        \hline
    \end{tabular}
    \label{tab:my_label1}
        \vspace{-0.5cm}
\end{table}

\section{Comparative Study with different reward functions}
 We conducted experiments with different possible reward designs as shown in Table~\ref{tab:my_label4}. Empirically, we found that combining rewards from both ranking and feature embedding space through triplet loss offers most optimum performance.

\setlength{\tabcolsep}{4.5pt}
\begin{table}[!hbt]
    \centering
    \caption{Performance analysis using different reward designs.}
    \footnotesize
    \begin{tabular}{ccccccc}
        \hline
         &  &  & \multicolumn{2}{c}{Chair-V2} & \multicolumn{2}{c}{Shoe-V2} \\
        \cline{4-7}
         &  \multicolumn{2}{c}{Rewards}  & Acc@1 & Acc@5 & Acc@1 & Acc@5\\
        \hline
         & \multicolumn{2}{c}{-rank} & 63.5 \% & 78.6\% & 42.6\% & 73.7\%\vspace{0.08cm} \\ 
         & \multicolumn{2}{c}{$\frac{1}{rank}$} & 64.2\% & 78.8\% & 43.2\% & 74.2\% \vspace{0.08cm}\\
        & \multicolumn{2}{c}{-$\mathcal{L}_{triplet}$} & 62.4\% & 78.1\% & 41.6\% & 72.7\% \vspace{0.08cm}\\ 
         & \multicolumn{2}{c}{$\frac{1}{\mathcal{L}_{triplet}  + \epsilon}$} & 60.2\% & 77.3\% & 38.8\% & 68.6\% \vspace{0.08cm}\\
          & \multicolumn{2}{c}{$\frac{1}{rank} -\mathcal{L}_{triplet}$}  & 64.8\% & 79.1\% & 43.7\% & 74.9\% \vspace{0.08cm}\\
        \hline
    \end{tabular}
    \label{tab:my_label4}
     \vspace{-0.3cm} 
\end{table}

 \vspace{-0.1cm}
\begin{figure}[!hbt]
\centering
\includegraphics[width=0.8\linewidth]{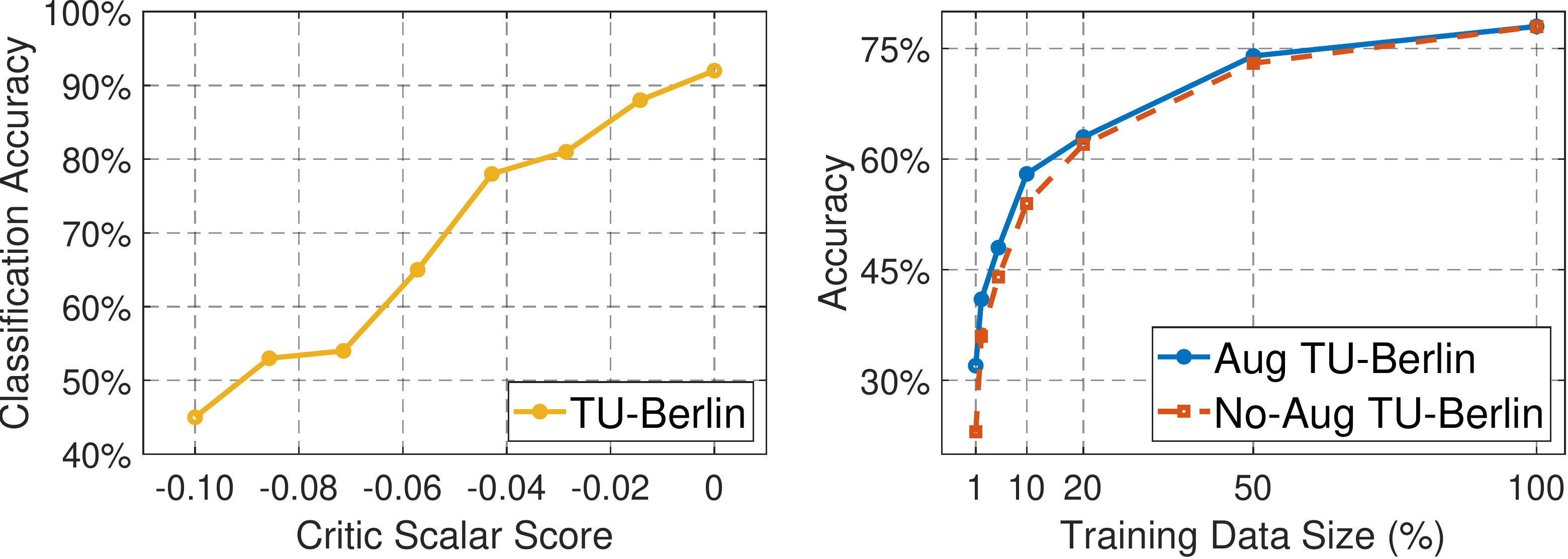}
\caption{(a) Retrieval ability of partial sketch: correlation between critic network $\mathrm{V(S)}$ predicted score and ranking percentile (b) Performance at varying training data size with stroke-subset selector based data augmentation.}
\label{fig_partial}
\vspace{-0.2cm}
\end{figure} 

\section{Classification Ability and Data Augmentation for sketch classification}
Similar to fine-grained retrieval \cite{bhunia2020sketch, bhunia2021more}, we extend our RL-based stroke-subset selector framework for classification task to judge if the critic network could be used to judge the recognition potential from partial sketch. To this end, we use negative of cross-entropy loss as the reward to train the stroke-subset selector under a pre-trained sketch classification network (Resnet50) on TU-Berlin dataset \cite{bhunia2021vectorization}. We obtain a similar correlation between critic network predicted score and classification  accuracy as shown in Fig.~\ref{fig_partial}.  In brief, the samples having higher scalar score predicted by the critic network tends to have a higher classification accuracy, thus proving the efficiency of modelling recognition ability of sketches through our framework. 

Similarly, one can use the stroke-subset selector (policy network) to augment the sketches for classification problem. Performance at varying training data regime is shown in Fig.~\ref{fig_partial} on TU-Berlin dataset.

\section{Motivation on removing ``fear'' for sketching}
 By removing ``fear'', we meant injecting that extra confidence to the users, knowing that even if they can not sketch well, the system will still be able to return favourable results.

\section{What happens with extreme cases?}

\begin{figure}[!hbt]
\centering
\includegraphics[height=3.5cm,width=0.6\linewidth]{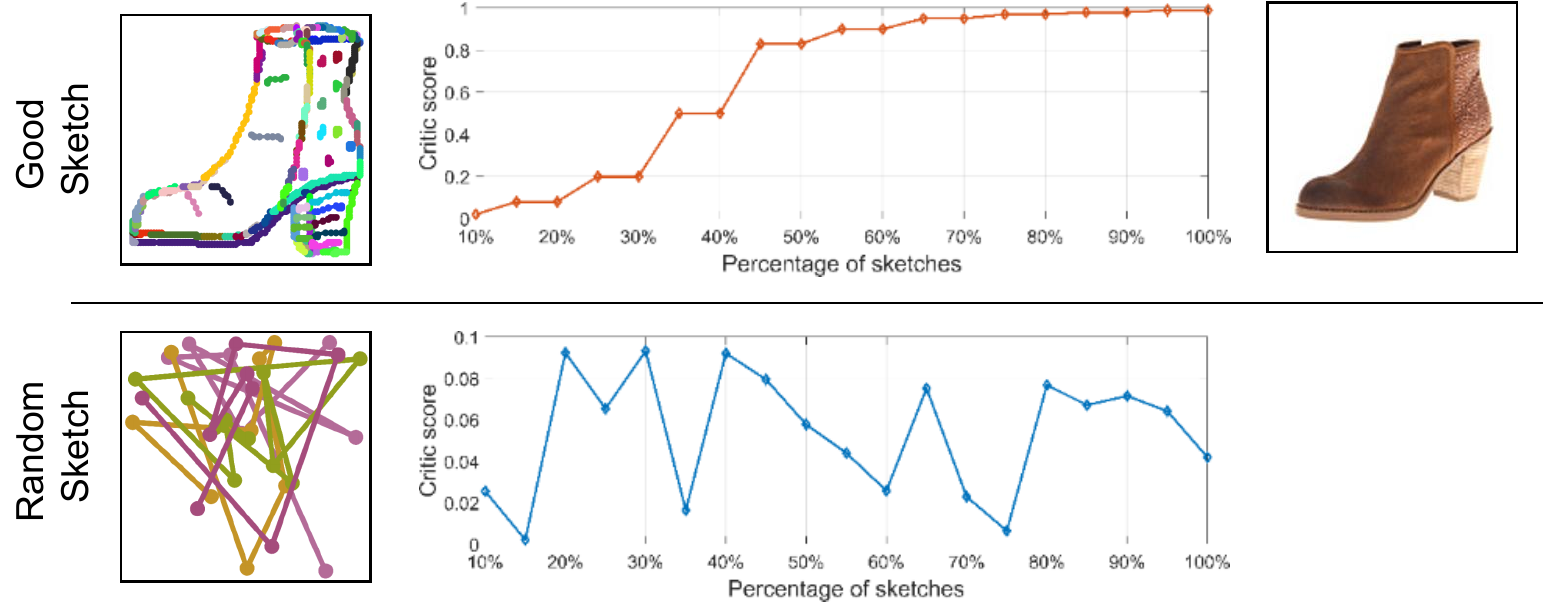}
\vspace{-0.5cm}
\caption{Critic score at progressive sketch drawing episode.}
\label{figextreme}
\end{figure}

The extreme case of completely random junk can be handled by our critic network, which will assign a low retrieval ability score, helping us to sidestep such unusable instances. On the other hand, critic network assigns progressively higher score for sketches from professional artists, and achieves retrieval threshold much earlier. Fig.~\ref{figextreme} offers examples of how the critic score changes for a good/professional sketch (top) and a complete random one (bottom).

\section{Clarity of binary stroke selection scheme}
In our framework, we have modelled the stroke selection through categorical distribution (softmax normalisation over $\mathbb{R}^2$). However, as the reviewer suggested, one could model using Bernoulli distribution where the stroke selector would predict a single sigmoid normalised scalar value $\mathbb{R}^1$. We tried both approaches and empirically found the use of categorical distribution to be more stable with faster convergence and quantitatively better results (by $1.41\%$ Acc@1 on ShoeV2). We will specifically mention this in the supplementary with a thorough ablative study upon acceptance.

\section{Training-time comparison with baselines}  
For the baselines, we \emph{do not} augment the sketches using \emph{all possible} stroke-subset combinations (with cost $\mathcal{O}(2^N)$). Taking into account all possible stroke subsets not only slows down the training data-pipeline, but many of these augmented sketch subsets are too coarse/incomplete to convey any useful information about the paired photo. Some initial experiments indicated model collapse due to noisy gradients raised from such overly coarse/incomplete sketch-subsets. Therefore, in order to eliminate noisy gradients in the baselines, we drop strokes at random while ensuring that the percentage of sketch vector length never falls below a certain threshold --  $80\%$ was empirically found to yield optimum performance.

To ensure fair comparisons, we also keep each model training until we find no further improvement in both the loss value and accuracy on the validation set for consecutive 20K iterations. Furthermore, under our experimental setup, the training time for all baselines as well as our method lies between 12-14 hours, ensuring a largely uniform training time for all.

\section{ Clarity on Training dataset} 
We use 6051+1800  images to train both the retrieval model and stroke-subset selector. In particular, first, we pre-train the retrieval model on raster sketches. Next, we use the sketch-vector modality of the same set of sketches to train the stroke-subset selector. It should be noted that while the retrieval model trained from raster sketches is unaware of stroke-specific importance for retrieval, the stroke-subset selector intelligently manages to eliminate the noise/inconsistent sketch strokes. Testing is done on the remaining 679+200 images which are never used in either stage of the training. 

\section{Comparison with soft-attention}In order to deal with partial sketches, one alternative is indeed to apply soft-spatial attention in raster-space, as used in Triplet-Attn-HOLEF \cite{song2017deep}. Through fusing Triplet-Attn-HOLEF with our Augment baseline, we devise a new baseline Triplet-Attn-HOLEF+Augment, which is able to achieve Acc@1(Acc@5) of $34.6\%(68.9\%)$ on the ShoeV2 dataset. {This is slightly better than the Augment baseline but significantly falls behind our final results. This further verifies the necessity of our stroke-subset selector to deal with the erroneous/noisy strokes that are inherent to the drawing process.}

\section{Limitations}  Cross-dataset generalisation for the stroke-subset selector in particular is an intriguing research direction, which we intend to cover in the future. Also, replacing the non-differentiable rasterization operation (sketch-vector to sketch-image) with a differentiable approximated one would be an interesting direction to explore too. This would make the whole pipeline end-to-end differentiable, so it can backpropagate the gradient calculated from triplet loss directly onto the stroke-subset selector without needing any RL-based formulation, ultimately increasing stability and pace of training.
  

\end{document}